\documentclass[lettersize,journal]{IEEEtran}
\usepackage{amsmath,amsfonts}
\usepackage{algorithmic}
\usepackage{algorithm}
\usepackage{array}
\usepackage[caption=false,font=normalsize,labelfont=sf,textfont=sf]{subfig}
\usepackage{textcomp}
\usepackage{stfloats}
\usepackage{url}
\usepackage{verbatim}
\usepackage{graphicx}
\usepackage{cite}
\usepackage{multirow}  
\usepackage{booktabs}  
\usepackage{url}
\usepackage[breaklinks,colorlinks,allcolors=blue]{hyperref}
\newcommand{\red}[1]{\textcolor{red}{#1}}
\newcommand{\blue}[1]{\textcolor{blue}{#1}}

\usepackage{xcolor}	
\usepackage{colortbl}	
\usepackage{cleveref} 

\begin{document}

\title{Real-World Remote Sensing Image Dehazing: Benchmark and Baseline}

\author{Zeng-Hui Zhu$^\dagger$, Wei Lu$^\dagger$, Si-Bao Chen$^\ast$, Chris H. Q. Ding, Jin Tang, and Bin Luo
	\thanks{$^\dagger$Equal contribution: Zeng-Hui Zhu and Wei Lu. $^\ast$Corresponding author: Si-Bao Chen.
		This work was supported in part by NSFC Key Project of International (Regional) Cooperation and Exchanges (No. 61860206004), NSFC Key Project of Joint Fund for Enterprise Innovation and Development (No. U20B2068, U24A20342) and National Natural Science Foundation of China (No. 61976004).}
	\thanks{Zeng-Hui Zhu, Wei Lu, Si-Bao Chen, Jin Tang, and Bin Luo are with the MOE Key Laboratory of ICSP, IMIS Laboratory of Anhui, Anhui Provincial Key Laboratory of Multimodal Cognitive Computation, Zenmorn-AHU AI Joint Laboratory, School of Computer Science and Technology, Anhui University, Hefei 230601, China (e-mail:1525074487@qq.com; luwei\_ahu@qq.com; sbchen@ahu.edu.cn; tangjin@ahu.edu.cn; luobin@ahu.edu.cn).}
	\thanks{Chris H. Q. Ding is with the School of Data Science (SDS), Chinese University of Hong Kong, Shenzhen 518172, China (e-mail: chrisding@cuhk.edu.cn).}}

\markboth{IEEE TRANSACTIONS ON GEOSCIENCE AND REMOTE SENSING, 2025}%
{Shell \MakeLowercase{\textit{et al.}}: A Sample Article Using IEEEtran.cls for IEEE Journals}

\maketitle

\begin{abstract}
Remote Sensing Image Dehazing (RSID) poses significant challenges in real-world scenarios due to the complex atmospheric conditions and severe color distortions that degrade image quality. The scarcity of real-world remote sensing hazy image pairs has compelled existing methods to rely primarily on synthetic datasets. However, these methods struggle with real-world applications due to the inherent domain gap between synthetic and real data. To address this, we introduce Real-World Remote Sensing Hazy Image Dataset (RRSHID), the first large-scale dataset featuring real-world hazy and hazy-free image pairs across diverse atmospheric conditions. Based on this, we propose MCAF-Net, a novel framework tailored for real-world RSID. Its effectiveness arises from three innovative components: Multi-branch Feature Integration Block Aggregator (MFIBA), which enables robust feature extraction through cascaded integration blocks and parallel multi-branch processing; Color-Calibrated Self-Supervised Attention Module (CSAM), which mitigates complex color distortions via self-supervised learning and attention-guided refinement; and Multi-Scale Feature Adaptive Fusion Module (MFAFM), which integrates features effectively while preserving local details and global context. Extensive experiments validate that MCAF-Net demonstrates state-of-the-art performance in real-world RSID, while maintaining competitive performance on synthetic datasets. The introduction of RRSHID and MCAF-Net sets new benchmarks for real-world RSID research, advancing practical solutions for this complex task. The code and dataset are publicly available at 
\url{https://github.com/lwCVer/RRSHID}. 

\end{abstract}


\section{Introduction}\label{Introduction}
\IEEEPARstart{R}{emote} sensing (RS) imagery serves as a pivotal enabler for environmental sciences and agricultural applications, driving significant advancements in urban planning, agricultural monitoring, disaster management, and environmental conservation \cite{chi2023trinity,jiang2023dehazing,gao2023task,lu2023robust, lu2024decouplenet}. However, the ubiquitous presence of atmospheric haze presents a persistent obstacle, degrading image quality by diminishing contrast and obscuring fine details. Such degradation severely compromises the precision and reliability of subsequent analytical tasks, underscoring the urgent need for efficient, robust dehazing algorithms tailored to practical RS applications.

\begin{figure}[t] \centering
\includegraphics[width=1\linewidth]{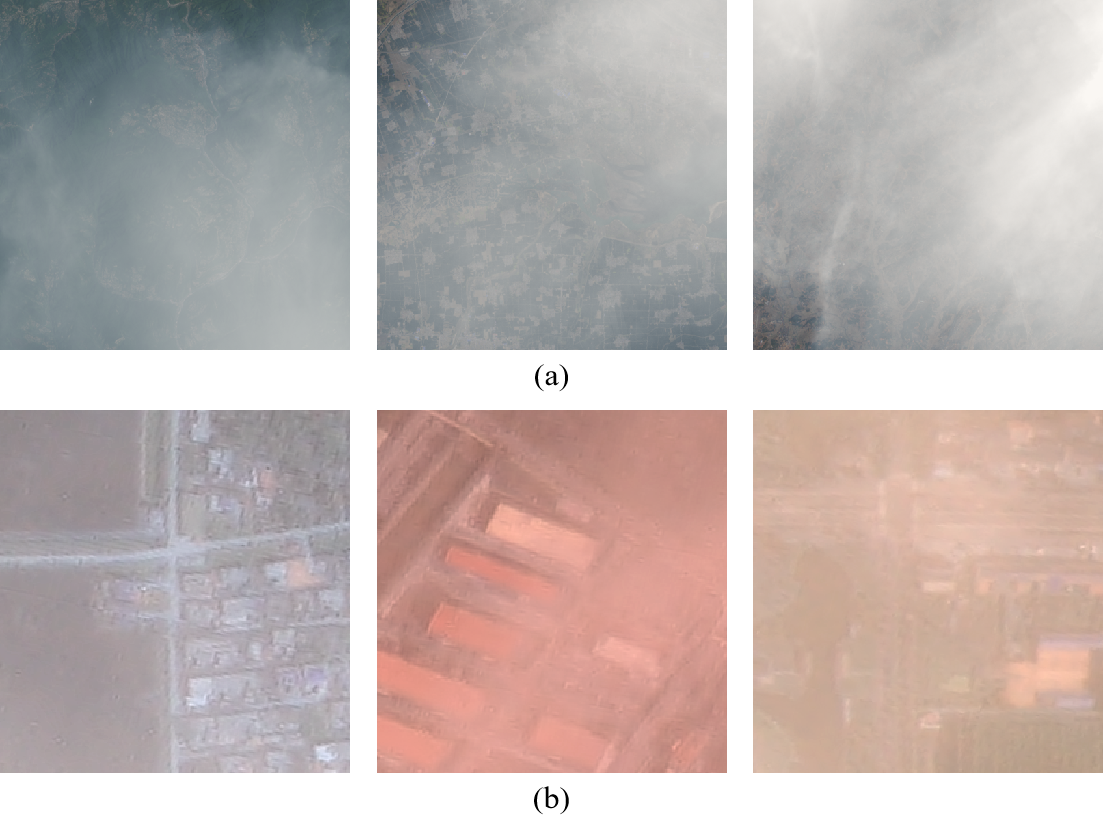}\vspace{-1mm}
\caption{Visual comparison of synthetic and real-world RS hazy images. (a) Synthetic hazy images from the RS-HAZE dataset \cite{song2023vision}. (b) Real-world hazy images from our RRSHID dataset, highlighting complex color variations.}	\label{fig:example}	\vspace{-2mm}
\end{figure}

In recent years, deep learning has catalyzed remarkable progress in RS image dehazing (RSID)  \cite{he2010single}. Pioneering efforts such as TransWeather \cite{Valanarasu_2022_CVPR} introduced Transformer architectures to dehazing, while 4KDehazing \cite{xiao2024single} advanced high-resolution dehazing through sophisticated multi-scale feature extraction. These methods have predominantly been developed and validated on synthetic datasets like RICE \cite{lin2019remote} and RS-HAZE \cite{song2023vision}, achieving impressive results—e.g., FFA-Net \cite{qin2020ffa} with a Peak Signal-to-Noise Ratio (PSNR) of 33.52 dB on RICE, and DehazeFormer \cite{song2023vision} with 39.57 dB on RS-HAZE. However, when deployed on real-world RS hazy images, these models frequently suffer substantial performance drops. This gap arises from three critical challenges:
\begin{enumerate}
	\item \textit{Domain Discrepancy:} Synthetic datasets rely on simplified atmospheric scattering models \cite{narasimhan2002vision}, which inadequately represent the intricate, non-linear atmospheric dynamics of real-world scenes.
	\item \textit{Complex Atmospheric Variability:} Real-world RS imagery contends with altitude-dependent atmospheric effects and spatially heterogeneous haze distributions \cite{shao2020domain}, which vary with elevation, time, and geography—conditions significantly more complex than those in synthetic data \cite{song2023vision}.
	\item \textit{Color Distortion Challenges:} Real-world hazy images display pronounced color distortions caused by multiple scattering and sensor-specific responses \cite{qin2018dehazing}, as shown in Fig. \ref{fig:example}, in stark contrast to the basic color shifts observed in synthetic datasets.
\end{enumerate}

To address these challenges, we introduce the Real-World Remote Sensing Hazy Image Dataset (RRSHID). It is the first large-scale dataset featuring real-world hazy and hazy-free image pairs across diverse atmospheric conditions and geographical contexts. Its construction is detailed in \Cref{dataset}. 

With this dataset established, we propose the Multi-branch Color-calibrated Adaptive Fusion Network (MCAF-Net), which demonstrates state-of-the-art (SOTA) performance in real-world RSID challenges.

At its core, MCAF-Net utilizes the Multi-branch Feature Integration Block Aggregator (MFIBA) as its primary encoding component, consisting of three cascaded Multi-branch Feature Integration Blocks (MFIBs). Each MFIB splits input features into four parallel branches—height-width (hw), channel-height (ch), channel-width (cw), and point-wise integration—using a channel-splitting strategy \cite{zhang2020multi}. Tailored convolution and adaptive parameters within each branch extract multi-scale, multi-dimensional feature representations, with grouped convolutions and parameter sharing ensuring computational efficiency. By cascading three MFIBs, MFIBA enhances feature robustness and regularizes information flow, adeptly balancing efficiency and representational capacity for real-world RSID.

Traditional RSID methods often adopt simplistic reconstruction techniques, such as direct feature mapping \cite{zhang2017beyond} or basic skip connections \cite{mao2016image}, neglecting the pivotal role of color calibration in real-world contexts. To address this, we introduce the Color-calibrated Self-supervised Attention Module (CSAM). CSAM combines self-supervised color learning with attention-driven refinement via two mechanisms: (1) an intermediate "fake image" generated through a learnable color correction matrix for implicit color supervision, and (2) a query-key-value attention mechanism where color-calibrated features guide refinement through a global channel attention branch (for broad color consistency, e.g., sky regions) and a local depth-wise convolution branch (for detail preservation, e.g., urban edges). Tailored for RSID and inspired by recent restoration advances \cite{zamir2022restormer}, CSAM bridges encoder and decoder, ensuring both color accuracy and structural integrity.

Moreover, conventional feature fusion in RSID—often limited to 1$\times$1 convolutions or selective kernel methods \cite{song2023vision}—struggles with the spatial heterogeneity and scale-varying degradation of real-world haze. These approaches either oversimplify feature interactions or rely on static receptive fields, failing to capture spectral-spatial complexity. Our Multi-scale Feature Adaptive Fusion Module (MFAFM) overcomes these shortcomings by integrating features across scales and layers via a three-branch architecture (3$\times$3, 5$\times$5, 7$\times$7 kernels). Dual attention—channel and spatial—prioritizes salient features, while residual learning and adaptive adjustments stabilize training and align features effectively. This design ensures adaptive, multi-scale fusion, preserving local details and global coherence critical for real-world RSID.

Our main contributions are summarized as follows:
\begin{enumerate}
    \item RRSHID as the first real-world RS hazy image dataset, highlighting how it addresses fundamental limitations in existing synthetic datasets.
    \item MCAF-Net specifically tailored to the challenges of real-world RSID, rather than simply adapting general image dehazing techniques.
    \item Our proposed modules (MFIBA, CSAM, and MFAFM) that specifically address the unique characteristics of real-world RS scenarios, including complex haze distributions, color distortion issues, and the need to preserve both fine details and global context.
    \item The practical significance of our approach for real-world RS applications, emphasizing computational efficiency and performance improvements in authentic scenarios.
\end{enumerate}
\section{Related Work}	\label{Related}

\subsection{Model-Based RSID}

Model-based methods for RSID have historically leveraged physical models of atmospheric scattering to estimate and mitigate haze effects. Pioneering efforts, such as the Dark Channel Prior (DCP) \cite{he2010single}, relied on the assumption that haze-free images contain dark pixels to derive transmission maps. However, this assumption often falters in RS scenarios due to expansive bright regions, such as deserts or snow-covered landscapes. To address this, subsequent innovations like the Haze-Lines Prior \cite{berman2018single} exploited linear pixel relationships in hazy images, while the Color Attenuation Prior \cite{zhu2015fast} utilized depth-color attenuation correlations to refine estimates. Despite these advancements, such methods frequently struggle to accommodate the diverse atmospheric conditions and intricate scene compositions prevalent in RS imagery. More recent developments, exemplified by Non-Local Image Dehazing \cite{berman2016non}, have integrated spatial context and global information to bolster performance. Nevertheless, model-based approaches remain challenged by varying atmospheric dynamics and the preservation of fine details in RS images.

\subsection{Deep Learning-Based RSID}

The advent of deep learning has transformed RSID, surpassing the limitations of traditional model-based techniques. Early contributions, such as DehazeNet \cite{cai2016dehazenet} and MSCNN \cite{ren2016single}, pioneered convolutional neural networks (CNNs) for transmission map estimation, establishing a benchmark for superior performance. AOD-Net \cite{li2017aod} advanced this paradigm by introducing end-to-end training for clean image recovery. As the field matured, attention mechanisms and multi-scale architectures gained traction. FFA-Net \cite{qin2020ffa} and GridDehazeNet \cite{liu2019griddehazenet} incorporated these features to effectively capture global context and local intricacies. RSDehazeNet \cite{guo2020rsdehazenet} and HyperDehazing \cite{fu2024hyperdehazing} introduced attention mechanisms specifically designed for multispectral and hyperspectral RS images. PSMB-Net \cite{sun2023partial} extended this trajectory with dual-path feature interactions to tackle non-uniform haze in urban RS scenes, while DR3DF-Net \cite{sun2025dynamic} enhanced feature integration through dynamic routing, enabling adaptive fusion across scales.

Recent innovations have shifted toward transformer-based architectures, marking a significant evolution in RSID. DehazeFormer \cite{song2023vision} leveraged shifted window attention to model long-range dependencies, offering robust haze removal across complex scenes. RSDformer \cite{song2023learning} further enhanced image structure recovery by incorporating novel self-attention mechanisms to capture both local and global correlations. PCSformer \cite{zhang2024proxy} built upon this by introducing cross-stripe attention modules to capture vertical and horizontal haze gradients, augmented by proxy tokens for computational efficiency. Trinity-Net \cite{chi2023trinity} integrated gradient-guided Swin transformers with physics-informed priors, facilitating joint optimization of dehazing and downstream tasks, such as land cover classification. Concurrently, state space models have emerged as compelling alternatives to traditional CNNs and transformer architectures. Notably, RSDehamba \cite{zhou2024rsdehamba} and HDMba \cite{fu2024hdmba} pioneered the application of the Mamba architecture to RSID and hyperspectral RSID, respectively, demonstrating superior efficiency in modeling long-range dependencies with linear computational complexity.

To overcome data scarcity—a persistent challenge in RSID—hybrid learning strategies have emerged. SCANet \cite{guo2023scanet} proposed a self-paced semi-curricular attention network, blending curriculum learning with channel-spatial attention to address non-homogeneous haze. Semi-supervised methods \cite{li2020semi} harnessed both labeled and unlabeled data, while domain adaptation techniques \cite{shao2020domain} bridged synthetic and real-world domains. Self-supervised learning \cite{liang2025image} has also shown promise in exploiting unlabeled RS data for improved dehazing. Recent unpaired training approaches have effectively addressed the synthetic-real domain gap. Diff-Dehazer \cite{lan2025exploiting} integrates diffusion priors with CycleGAN for bijective mapping, leveraging both physical priors and cross-modal representations. DNMGDT \cite{su2025real} utilizes a parameter-shared architecture with adaptive weighting of prior-based pseudo labels and physical model guided domain transfer to enhance generalization to real hazy scenes. SFSNiD \cite{cong2024semi} explored frequency-consistent twin fusion, while MaxDehazeNet \cite{wu2023maxdehazenet} applied maximum flow theory to model atmospheric scattering, enhancing physical interpretability.

\begin{figure*}[t]
	\centering
	\includegraphics[width=1\textwidth]{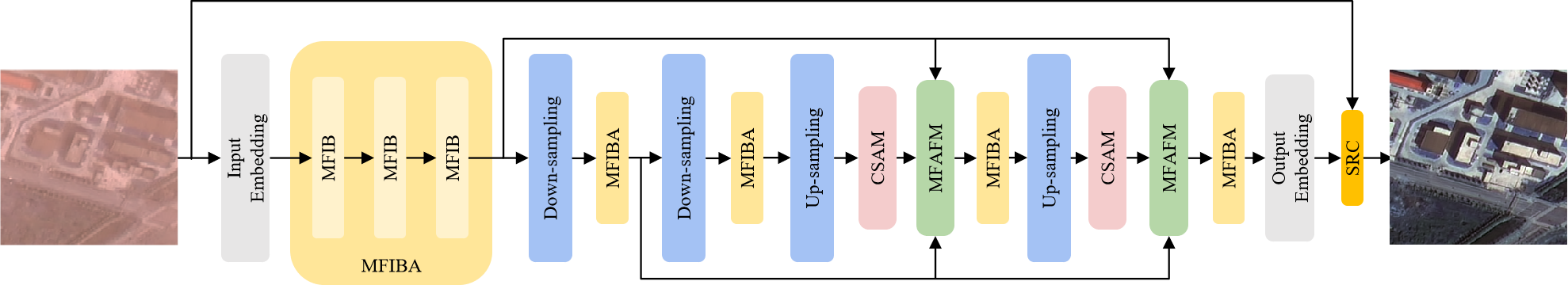}\vspace{-1mm}
	\caption{The proposed MCAF-Net architecture. It adopts a U-shaped design for real-world RSID. The encoder utilizes the MFIBA with three cascaded MFIBs, while the decoder integrates CSAM and MFAFM, creating a robust framework for real-world dehazing.}
	\label{fig: MCAF-Net}
	\vspace{-2mm}
\end{figure*}

\section{Proposed Method} \label{sec:method}

In this section, we detail the architecture and key components of the proposed MCAF-Net, specifically engineered for RSID. As illustrated in Fig. \ref{fig: MCAF-Net}, MCAF-Net employs a U-shaped architecture with five stages, integrating three novel modules: MFIBA, CSAM and MFAFM.

The encoder leverages MFIBA to perform efficient, cross-dimensional feature extraction through cascaded multi-branch processing, addressing the spatial heterogeneity of real-world haze. 
CSAM bridges the encoder and decoder by progressively refining color distortion through attention-guided refinement, ensuring structural preservation. In the decoder, MFAFM adaptively fuses multi-scale features using dual attention mechanisms—channel attention for color channel recalibration and spatial attention for terrain-specific haze suppression—effectively balancing local details and global context. This design is tailored to tackle the complex atmospheric variability and color distortions prevalent in real-world RSID scenarios, outperforming conventional approaches reliant on static feature processing.

\begin{figure}[t] \centering
	\includegraphics[width=1\linewidth]{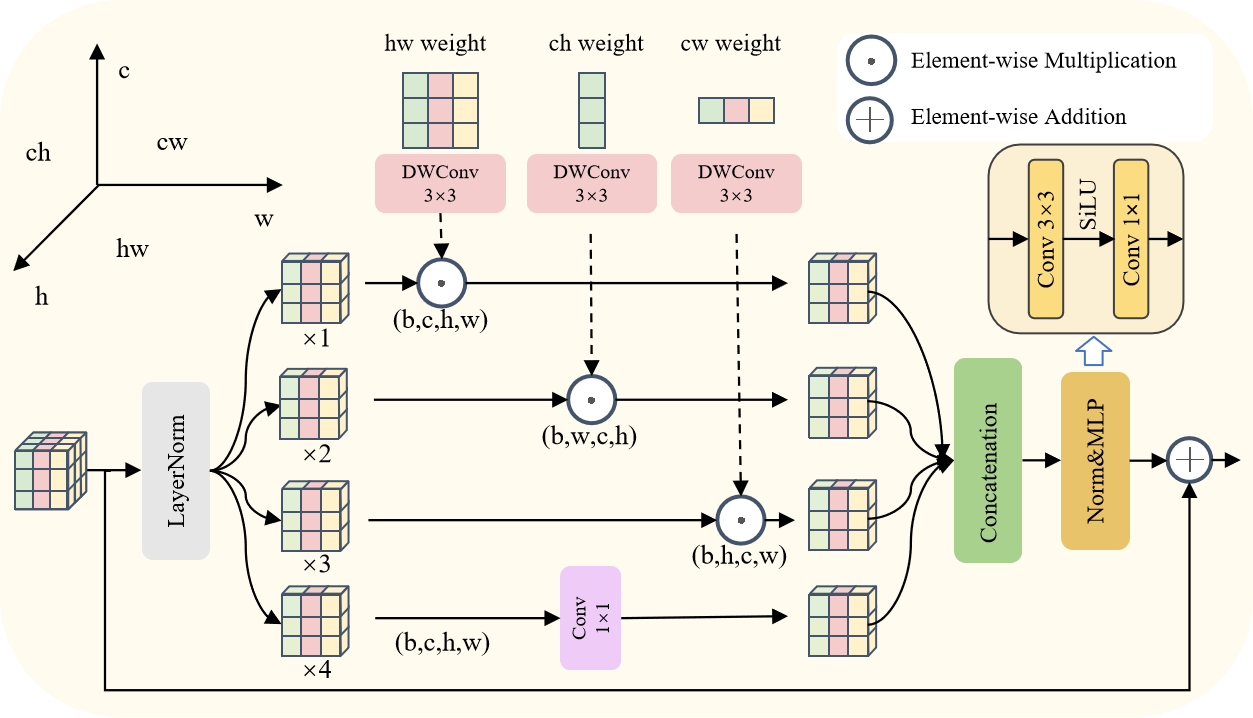}\vspace{-1mm}
	\caption{Structure of the proposed MFIB. It employs axis-oriented attention to refine feature representation and flow, achieving superior dehazing performance with reduced computational overhead.}	\label{fig: MFIB}
	\vspace{-2mm}
\end{figure}

\subsection{MFIBA}
Conventional RSID methods uniformly process features across all dimensions, propagating redundant information that obscures critical structural details and degrades generalization in complex atmospheric conditions. Real-world RSID requires robust feature extraction capable of addressing spatially heterogeneous haze distributions, a challenge unmet by conventional methods reliant on static full-channel convolutions \cite{li2019benchmarking, qin2020ffa}. To overcome this limitation, we propose the MFIBA, a lightweight module grounded in dynamic feature disentanglement theory \cite{chen2021pre}. MFIBA selectively amplifies haze-invariant features while suppressing noise, achieving a balance between computational efficiency and representational capacity tailored for real-world RSID.

MFIBA processes input features $\mathbf{X} \in \mathbb{R}^{b \times c \times h \times w}$ by partitioning them into four subgroups $\{x_1, x_2, x_3, x_4\}$, each with dimensions $(1, c/4, h, w)$. Three learnable queries—$\mathbf{Q}_{hw}$, $\mathbf{Q}_{ch}$ and $\mathbf{Q}_{cw}$—are dynamically generated for the spatial ($h$-$w$), channel-height ($c$-$h$), and channel-width ($c$-$w$) dimensions. All queries are independently randomly initialized. In multi-branch partial queried learning, we transform ${x_1}$, ${x_2}$, and ${x_3}$ into formats $(b, c, h, w)$, $(b, w, c, h)$, and $(b, h, c, w)$, respectively. We then apply bilinear interpolation to adjust the last two dimensions of queries to match feature groups, and employ depth-wise convolutions to enhance query representations. These queries interact with reshaped feature groups via Hadamard products $\mathbf{Q}_{ij} \odot \mathbf{X}_{ij}$, selectively attenuating redundant haze patterns while preserving cross-dimensional correlations. For ${x_4}$, we solely apply 1$\times$1 convolution for channel-wise information interaction, achieving partial self-relevant feature representation. Subsequently, the four groups of enhanced features are concatenated and transformed through a simple feed-forward network. A cascaded architecture with three MFIBs stages progressively refines contextual representations through residual connections, expanding the effective receptive field to capture large-scale haze structures. Computational efficiency is achieved through depth-wise separable convolutions \cite{howard2017mobilenets} and 1$\times$1 channel interactions, reducing complexity from $\mathcal{O}(c^2 h w k^2)$ to $\mathcal{O}(c^2 h w / 4 + 3c h w)$. This process is formalized as:
\begin{equation}
	\vspace{-1pt}
	\begin{split}
		&[X_{hw}, X_{ch}, X_{cw}, X_0] = \text{LN}(\mathbf{X}) \\
		& \mathbf{Q}_{hw}, \mathbf{Q}_{ch}, \mathbf{Q}_{cw} = \text{DWConv}_{3\times3}(\mathbf{Q}) \\
		&\mathbf{X}'_{queried} = \text{Concat}(\mathbf{Q}_{ij} \odot \mathbf{X}_{ij}, \text{Conv}_{1\times1}(X_0)) \\
		&\mathbf{X}_{queried} = \text{MLP}(\mathbf{X}'_{queried}) + \mathbf{X}
	\end{split}
	\vspace{-1pt}
\end{equation}
where $\text{LN}(\cdot)$ denotes layer normalization, $\text{DWConv}(\cdot)$ is depth-wise convolution and  $i, j \in \{hw, ch, cw\}$. The MFIBA further enhances contextual information by cascading multiple MFIBs. Its structure as follows:
\begin{equation}
	\vspace{-3pt}
	\text{MFIBA}(\mathbf{X})=\text{Conv}_{3\times3}(\text{MFIB}_3(\text{MFIB}_2(\text{MFIB}_1(\mathbf{X}))))
\end{equation}
By synergizing dynamic dimension-wise attention, multi-scale feature fusion, and lightweight operators, MFIBA establishes a theoretically grounded framework for efficient and adaptive dehazing, effectively addressing the spatiotemporal heterogeneity of real-world RS imagery without compromising computational practicality.


\subsection{CSAM}

\begin{figure}[t]	\centering
	\includegraphics[width=1\linewidth]{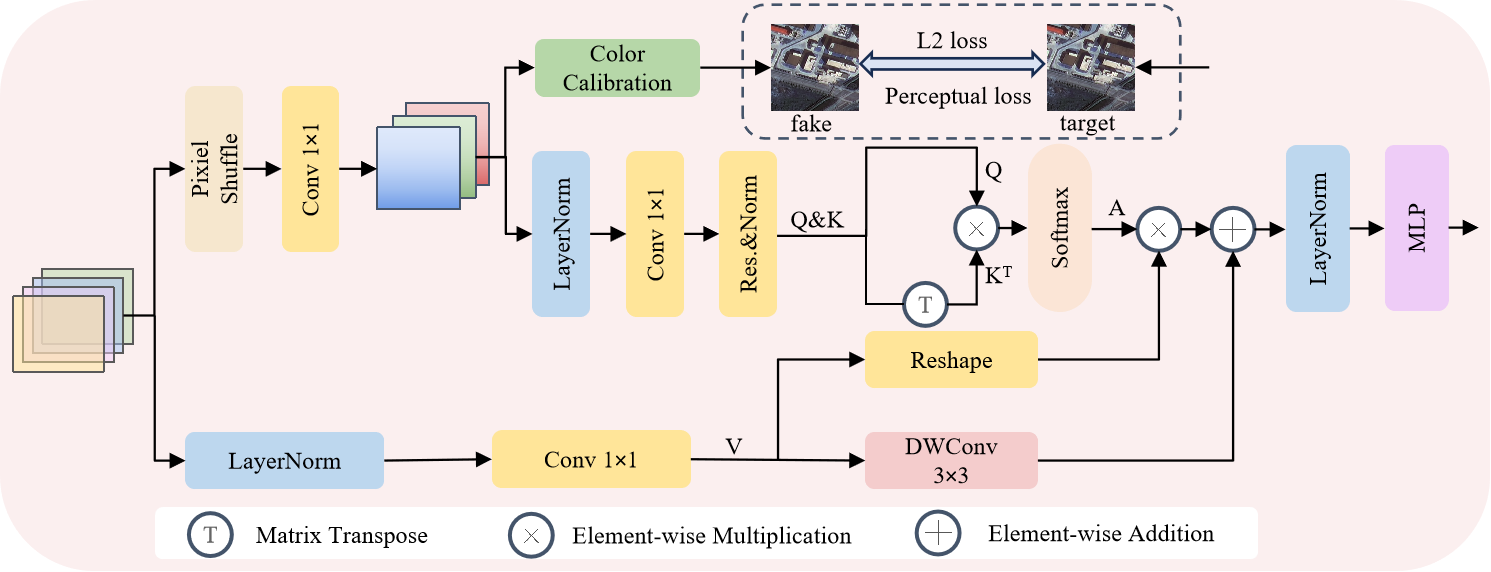}\vspace{-1mm}
	\caption{Architecture of CSAM. The color calibration branch (top) learns sensor-specific spectral adjustments via $W_{\text{color}}$, while the attention branch (bottom) integrates global channel dependencies and local spatial details via QKV processing.}
	\label{fig: CSAM}
	\vspace{-2mm}
\end{figure}

RS images under hazy conditions often exhibit severe color distortion, which is more complex compared to near-ground images, as atmospheric thickness and composition vary significantly at different altitudes. Conventional RSID \cite{li2022m2scn, li2023efficient, zheng2022dehaze} methods often overlook color calibration during early decoding stages \cite{liu2019single}, limiting their effectiveness against the severe color distortions \cite{qin2018dehazing} induced by atmospheric scattering in real-world scenarios. 
To address this, we introduce CSAM, integrated across multiple decoder levels, to progressively refine haze-free features while ensuring color accuracy and structural integrity, as shown in Fig. \ref{fig: CSAM}.

CSAM enables progressive refinement of haze-free features through three key operations. First, \textit{PixelShuffle} upsampling reconstructs high-resolution features, preserving channel-wise relationships. Second, a learnable color correction matrix \( W_{\text{color}} \in \mathbb{R}^{3 \times C} \) transforms features via \( I_{\text{fake}} = f_{\text{color}}(\text{Conv}_{1\times1}(X_{\text{up}})) \), where \( f_{\text{color}}(\cdot) \) implements spectral compensation through matrix multiplication and bias addition. This operation simulates the inverse process of atmospheric scattering by adaptively adjusting RGB channel gains and offsets \cite{yang2017rgb}. Third, the attention employs query-key-value (QKV) processing with depth-wise convolution refinement:
\begin{equation}
	\begin{split}
		\vspace{-3mm}
		X_{up} = U(X)&,  \hspace{4mm} Q, K = \text{Conv}_{1\times1}(\text{LN}(X_{up})) \\
		A &= \text{softmax}(\alpha \cdot \frac{QK^T}{\sqrt{d_k}}) \\
		F_c &= AV + \text{DWConv}_{3\times3}(V)
	\end{split}
\end{equation}
Here, $U(\cdot)$ denotes the upsampling operation. $\alpha$ is a learnable parameter and $d_k$ is the dimension of the key vectors. $V$ is the value vector and $\text{DWConv}_{3\times3}$ is a depth-wise separable convolution which preserves spatial details through parameter-efficient filtering, while the attention map \( A \in \mathbb{R}^{H \times W \times C} \) emphasizes haze-invariant channels (e.g., near-infrared bands less affected by scattering). This dual-stream design is biologically inspired by the parvocellular pathway for color processing and magnocellular pathway for spatial acuity in human vision \cite{ruan2022feature}.

Our proposed CSAM module offers two significant contributions. First, it operates specifically in the color space with a learnable correction matrix that adaptively adjusts color distribution based on input characteristics, addressing complex color distortion in haze-affected RS imagery. Second, CSAM implements an innovative self-attention mechanism that differs from similar strategies optimized for object tracking\cite{xue2023smalltrack,xue2024consistent} by: (1) focusing on color rather than spatial features to address differential atmospheric scattering effects across color channels; (2) maintaining multi-scale color consistency through MFIBA module integration; and (3) specifically targeting real-world RS dehazing challenges rather than general vision tasks. These design elements enable CSAM to effectively address color restoration challenges in RSID, achieving superior color correction while preserving critical image details.

\subsection{MFAFM}

\begin{figure}[t]
	\centering
	\includegraphics[width=1\linewidth]{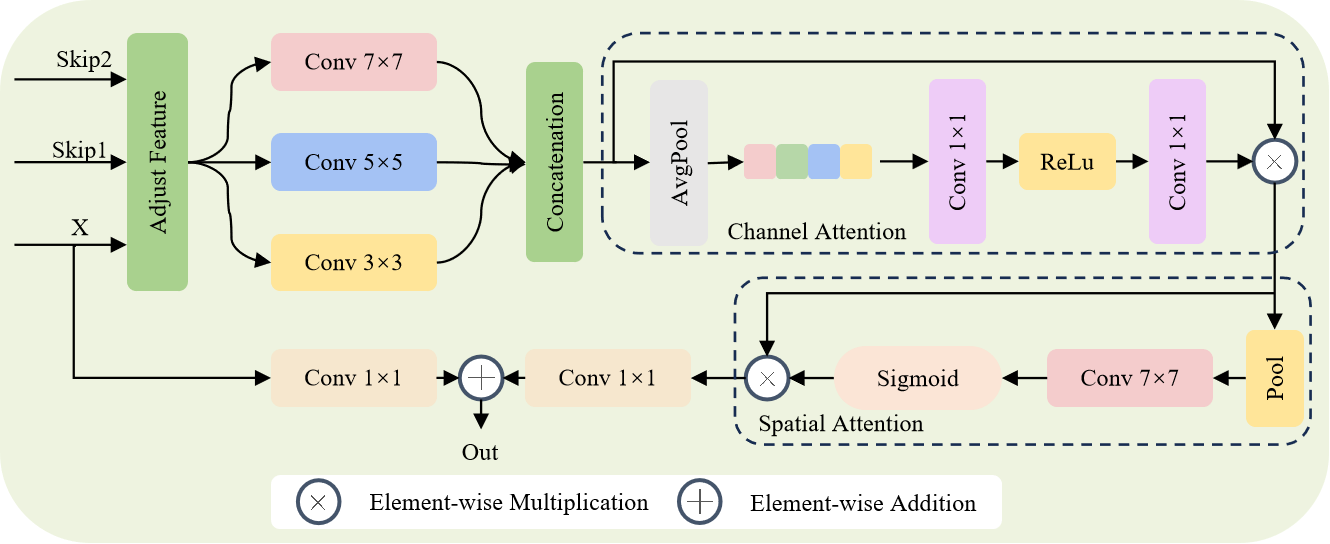}\vspace{-1mm}
	\caption{Architecture of MFAFM. Three convolutional branches capture scale-specific features, while dual attention mechanisms suppress haze-corrupted components. Residual connections preserve original feature integrity.}
	\label{fig:MFAFM}
	\vspace{-2mm}
\end{figure}

Traditional RSID feature fusion methods, such as 1$\times$1 convolutions and SKFusion \cite{song2023vision}, struggle to handle \textit{altitude-dependent haze variations}—where haze density increases exponentially at lower altitudes —and fail to adaptively balance fine details (e.g., urban edges) with the global atmospheric context. Fixed-scale operators cannot jointly address these spatially heterogeneous phenomena, leading to over-smoothing or residual artifacts. The proposed MFAFM module addresses the critical challenge of multi-scale feature fusion in real-world RSID, where altitude-dependent haze density variations and complex land cover patterns demand adaptive integration of spatial-contextual information across scales. As shown in Fig.~\ref{fig:MFAFM}, the module processes three inputs: the current feature map $x$ and two skip connections $\text{$skip1$}$, $\text{$skip2$}$ from preceding layers, each capturing distinct scale-specific characteristics. 

MFAFM employs convolutional kernels of sizes 3$\times$3, 5$\times$5, and 7$\times$7 to capture scale-specific haze patterns: small kernels preserve high-frequency details (e.g., building edges), medium kernels address mid-range gradients (e.g., vegetation clusters), and large kernels aggregate global context (e.g., sky-to-ground transitions). These operations are defined as:
\begin{equation}
	\begin{split}
		x_1 &= \text{Conv}_{3\times3}(x) \\
		x_2 &= \text{Conv}_{5\times5}(\text{$skip1$}) \\
		x_3 &= \text{Conv}_{7\times7}(\text{$skip2$})
	\end{split}
\end{equation}
The resulting feature maps are then concatenated. To enhance channel-wise feature relationships, we employ a channel attention mechanism \( \text{CA}(\cdot) \) that recalibrates RGB channel relationships to address color distortion. Spatial relationships, which identifies and processes regions with varying haze densities across different terrain types, are further refined using a spatial attention mechanism \( \text{SA}(\cdot) \):
\begin{equation}
	\begin{split}
		\    \text{X}_{\text{cat}} &= \text{Concat}(x_1, x_2, x_3) \\
		\text{X}_{\text{ca}} &= \text{X} \odot \text{CA}(\text{X}_{\text{cat}}) \\
		\text{X}_{\text{sa}} &= \text{X}_{\text{ca}} \odot \text{SA}(\text{X}_{\text{ca}})
	\end{split}
\end{equation}
The refined features are then fused with the original input through residual connections. MFAFM can be summarized as:
\begin{equation}
	\text{MFAFM}(x, \text{$skip1$}, \text{$skip2$}) = \text{Conv}_{1\times1}(\text{X}_{\text{sa}}) + \text{Conv}_{1\times1}(x)
\end{equation}
ensuring stable gradient flow while preserving critical high-resolution details. This hierarchical fusion strategy emulates the human visual system's multi-scale processing mechanism, where retinal ganglion cells progressively integrate local contrast and global luminance information.

\subsection{Loss Function}
MCAF-Net is optimized using a hybrid loss function combining $L_2$ and perceptual losses. While $L_2$ loss aligns predictions with ground truth for high PSNR, it may weaken structural similarity (SSIM). The perceptual loss, extracted from the 3rd, 8th, and 15th layers of a pretrained VGG16 network, enhances structural similarity and visual quality:
\begin{equation}
	L_{\text{perceptual}}(\tilde{J}, J) = \sum_{l \in \{3, 8, 15\}} L_2(\text{VGG}_l(\tilde{J}), \text{VGG}_l(J))
\end{equation}
Perceptual loss aids in preserving the structural and detail information of the images, thereby improving the visual quality of the dehazed images. The final loss function combines these two losses, expressed as:
\begin{equation}
	L(\tilde{J}, J) = L_2(\tilde{J}, J) + \lambda L_{\text{perceptual}}(\tilde{J}, J)  
\end{equation}
where $\lambda = 0.04$ balances the two components \cite{Valanarasu_2022_CVPR, gao2023frequency}.

\section{RRSHID: A Real-World RSID Dataset} \label{dataset}

RSID is a persistent challenge, particularly in real-world contexts where atmospheric conditions exhibit significant complexity and variability. While deep learning has driven notable progress in this domain, its application to authentic RS imagery remains constrained. Studies by Li \textit{et al.}~\cite{li2019benchmarking} and Qin \textit{et al.}~\cite{qin2020ffa} demonstrate that existing methods often falter under diverse atmospheric scattering patterns and heterogeneous scene compositions. Furthermore, Zhang \textit{et al.}~\cite{zhang2020deep} underscore the limited generalization of many models to real-world RS data, attributing this to the reliance on synthetic training datasets that fail to capture the intricacies of natural haze. To bridge this critical gap, we present the Real-World Remote Sensing Hazy Image Dataset (RRSHID), a pioneering collection of 3,053 paired hazy and haze-free images. spanning urban, agricultural, and coastal landscapes across China, including provinces such as Anhui, Sichuan, Shanghai, Shandong, and Fujian, these images were captured under diverse illumination and seasonal conditions. Unlike synthetic datasets such as those in~\cite{lin2019remote,song2023vision,huang2020single,chi2023trinity}, which depend on simplified atmospheric models, RRSHID leverages collaborative data acquisition with meteorological agencies. RRSHID ensures realistic representations of altitude-dependent haze distributions and sensor-induced color distortions. It serves as a vital benchmark for evaluating dehazing algorithms and advancing domain generalization in real-world RSID.

\subsection{Image Acquisition and Processing}
The development of RRSHID involved a systematic methodology for image selection and preprocessing, implemented through collaboration with remote sensing domain experts. Working in concert with meteorological agencies and geospatial scientists, we employed multi-temporal satellite image alignment techniques to identify spatially congruent regions captured by satellite platforms across multiple acquisition cycles. To ensure the authenticity and precision of our paired images, we developed a comprehensive dataset generation framework consisting of three key stages: 

\subsubsection{Data Source Selection}
 The dataset draws from the GF\_PMS: L1A optical satellite, part of the Optical and SAR Satellite Payload Retrieval system, which delivers high-resolution imagery at 1-meter panchromatic and 4-meter multispectral scales. Covering diverse terrains across China—urban centers, agricultural fields, and coastal zones—the images were acquired 2021-2023, reflecting seasonal and illumination variability. Each raw satellite image, stored in GeoTIFF format with an approximate file size of 8 GB, contains rich metadata including acquisition time, sensor parameters, and precise geolocation information.

\subsubsection{Temporal-Spatial Alignment}
The critical challenge in creating paired hazy/haze-free images lies in ensuring precise alignment between degraded and clear images. Our approach addresses this by a dual temporal-spatial alignment strategy:

\begin{itemize}
\item \textit{Temporal Proximity Selection}: We first identified temporally proximate multispectral image pairs (typically within 1-3 months) from the same geographic region that exhibited contrasting atmospheric conditions—one with significant haze presence and one with clear visibility. This temporal proximity minimizes seasonal variations in vegetation, building structures, and other land features.

\item \textit{Geospatial Coordinate Matching}: For each identified image pair, we implemented precise geospatial alignment by extracting geographic coordinate metadata (latitude/longitude) from image headers and identifying overlapping regions with consistent land features but different atmospheric conditions. Through precise marking of the top-left and bottom-right geographic coordinates of target regions, we ensured exact spatial correspondence before applying coordinate-based cropping to extract identical geographic areas from both images.
\end{itemize}

\subsubsection{Image Preprocessing}
Following alignment, each image pair underwent a standardized preprocessing workflow:
\begin{itemize}
\item \textit{Channel Conversion}: Multispectral images with multiple bands were transformed into 3-channel RGB (red, green, blue) format using Adobe Photoshop, prioritizing bands optimal for visual analysis.

\item \textit{Format Optimization}: Custom Python scripts refined the cropped images, generating $256\times256$ pixels subimages and converting them from TIFF to PNG format to enhance compatibility with machine learning frameworks.
\end{itemize}

\begin{table*}[t] \centering \scriptsize
	\renewcommand{\arraystretch}{1.2}
\setlength{\tabcolsep}{5pt}
\caption{Comparison of RRSHID with Existing Remote Sensing Datasets\label{tab:dataset_comparison}}\vspace{-1mm}
\begin{tabular}{lcccccccc}
\toprule
Dataset & Size & Real/Synthetic & Paired & Resolution & Application & Color Bias & Image Type & Scene Diversity\\
\midrule
StateHaze1K~\cite{huang2020single} & 1,200 & Synthetic & Yes & 512$\times$512 & Dehazing & No & RGB & Single \\
RSID~\cite{chi2023trinity} & 1,000 & Synthetic & Yes & 256$\times$256 & Dehazing & No & RGB & Single \\
RS-HAZE~\cite{song2023vision} & 51,300 & Synthetic & Yes & 512$\times$512 & Dehazing & No & RGB & Single \\
LHID~\cite{zhang2022dense} & 31,017 & Synthetic & Yes & 512$\times$512 & Dehazing & No & RGB & Multiple \\
DHID~\cite{zhang2022dense} & 14,990 & Synthetic & Yes & 512$\times$512 & Dehazing & No & RGB & Single \\
HyperDehazing~\cite{fu2024hyperdehazing} & 2,000 & Synthetic & Yes & 512$\times$512 & Dehazing & No &  Hyperspectral & Multiple \\
HyperDehazing~\cite{fu2024hyperdehazing} & 70 	& Real & No & 512$\times$512 & Dehazing & No &  Hyperspectral & Multiple \\
RICE1~\cite{lin2019remote} & 500 & Real & Yes & 512$\times$512 & Dehazing & No & RGB & Single \\
RICE2~\cite{lin2019remote} & 736 & Real & Yes & 512$\times$512 & Cloud/Haze Removal & No & RGB & Single \\
HRSI~\cite{liu2024oriented} & 796 & Real & No & 512$\times$512 to 4000$\times$4000 & Dehazing (Object Detection)  & No & RGB & Single \\
\rowcolor{gray!30}\textbf{RRSHID} & 3,053 & Real & Yes & 256$\times$256 & Dehazing & Yes & RGB & Multiple \\
\bottomrule
\end{tabular}\vspace{-2mm}
\end{table*}

\begin{figure}[t]
	\centering
	\includegraphics[width=1 \columnwidth]{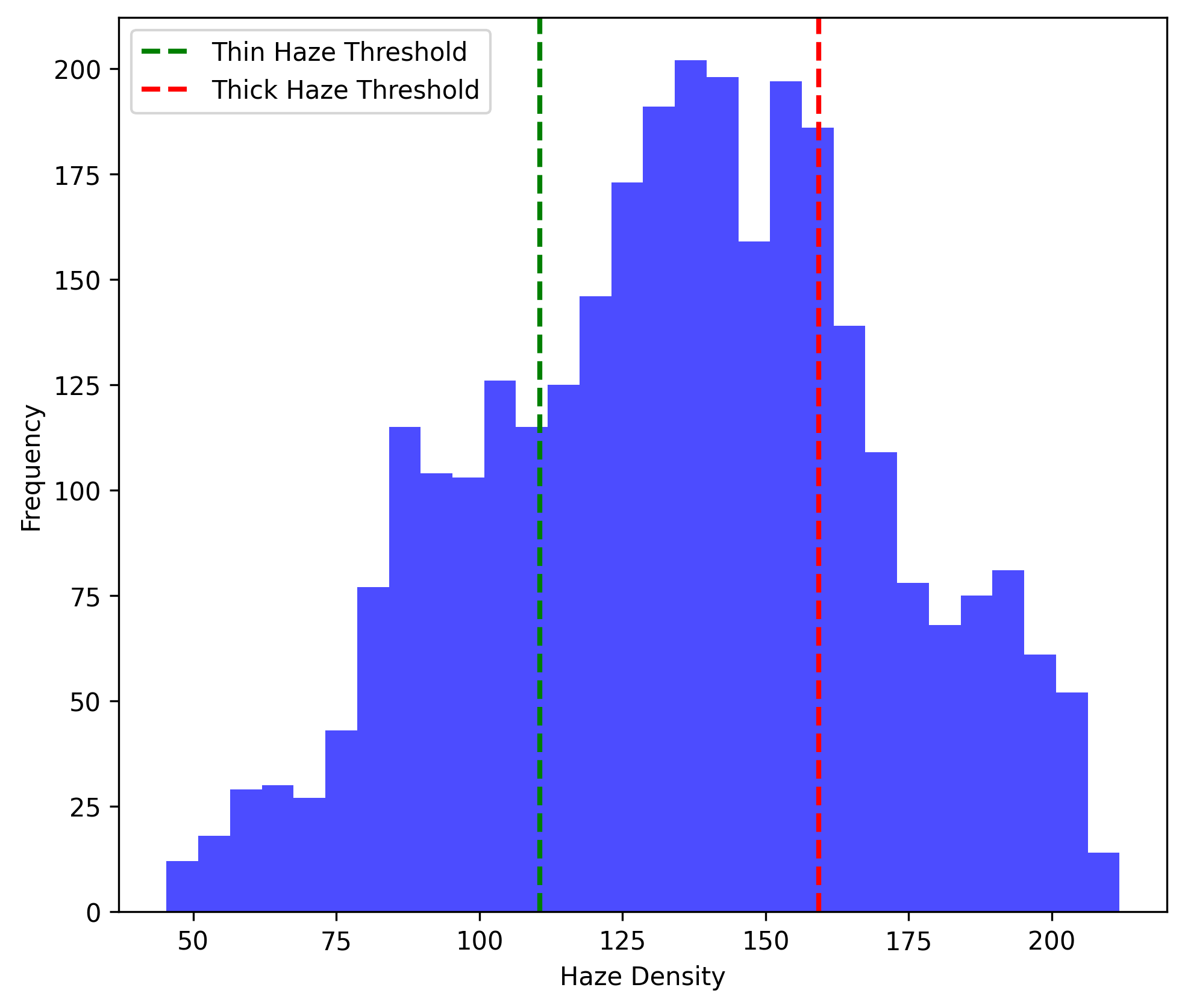}\vspace{-2mm}
	\caption{Haze density distribution across the RRSHID dataset.}
	\label{fig:hazy_density}
\end{figure}

\subsection{Haze Density Estimation}
To classify images by haze intensity, we adopted the dark channel prior method introduced by He \textit{et al.}~\cite{he2010single}. This technique estimates the transmission map $t(x)$ based on the principle that, in haze-free images, most non-sky regions exhibit at least one color channel with near-zero intensity. The dark channel is mathematically defined as:
\begin{equation}
	J_{\text{dark}}(x) = \min_{y\in\Omega(x)} \left( \min_{c\in\{r,g,b\}} J^c(y) \right),
\end{equation}
where $J^c$ denotes a color channel of the haze-free image $J$, and $\Omega(x)$ represents a local patch centered at pixel $x$. The transmission map is subsequently derived as:
\begin{equation}
	\tilde{t}(x) = 1 - \omega \min_{y\in\Omega(x)} \left( \min_{c} \frac{I^c(y)}{A^c} \right),
\end{equation}
where $I$ is the hazy image, $A$ is the atmospheric light, and $\omega$ is a constant (typically 0.95) to retain subtle haze effects. For each image, we calculated the mean dark channel value as a proxy for haze density. The threshold determination for haze categorization followed a data-driven approach. We performed k-means clustering (k=3) on the distribution of mean dark channel values across the entire dataset, which revealed natural groupings at 110.58 and 159.31. These thresholds were further validated through visual assessment by remote sensing experts and correlation with meteorological visibility data from ground stations (Pearson's r = 0.87), as depicted in Fig.~\ref{fig:hazy_density}. Images were accordingly categorized into thin (below 110.58), moderate (110.58–159.31), and thick (above 159.31) haze levels.


\subsection{Dataset Characteristics}

\begin{table}[!t] \centering \scriptsize
	\renewcommand{\arraystretch}{1.2}
	\setlength{\tabcolsep}{12pt}
	\caption{Distribution of Image Pairs in the RRSHID Dataset\label{tab:dataset_distribution}}\vspace{-1mm}
	\begin{tabular}{lrrrr}
		\toprule
		Haze Level & Total & Train & Test & Validation \\
		\midrule
		Thin haze         & 763    & 610   & 76    & 77   \\
		Moderate haze     & 1,526  & 1,220 & 152   & 154  \\
		Thick haze        & 764    & 611   & 76    & 77   \\
		\cmidrule{1-5}
		Total    & 3,053  & 2,441 & 304   & 308  \\
		\bottomrule
	\end{tabular}\vspace{-2mm}
\end{table}

RRSHID encompasses 3,053 pairs of hazy and haze-free RS images, constituting the largest real-world dataset of its kind, as comparison in Table~\ref{tab:dataset_comparison}. These pairs are stratified by haze density into three categories, as detailed in Table~\ref{tab:dataset_distribution}.

Each image, standardized at $256\times256$ pixels, strikes a balance between spatial detail and computational efficiency for deep learning applications. As the first large-scale, real-world paired RS dataset, RRSHID captures authentic atmospheric phenomena absent in synthetic counterparts, including:

· Heterogeneous haze densities and spatial distributions within individual images,

· Intricate interactions between haze and diverse land cover types (e.g., urban, vegetative, and marine),

· Color deviations caused by variations in natural lighting conditions and atmospheric composition.

\begin{figure}[t]
	\centering
	\includegraphics[width=\columnwidth]{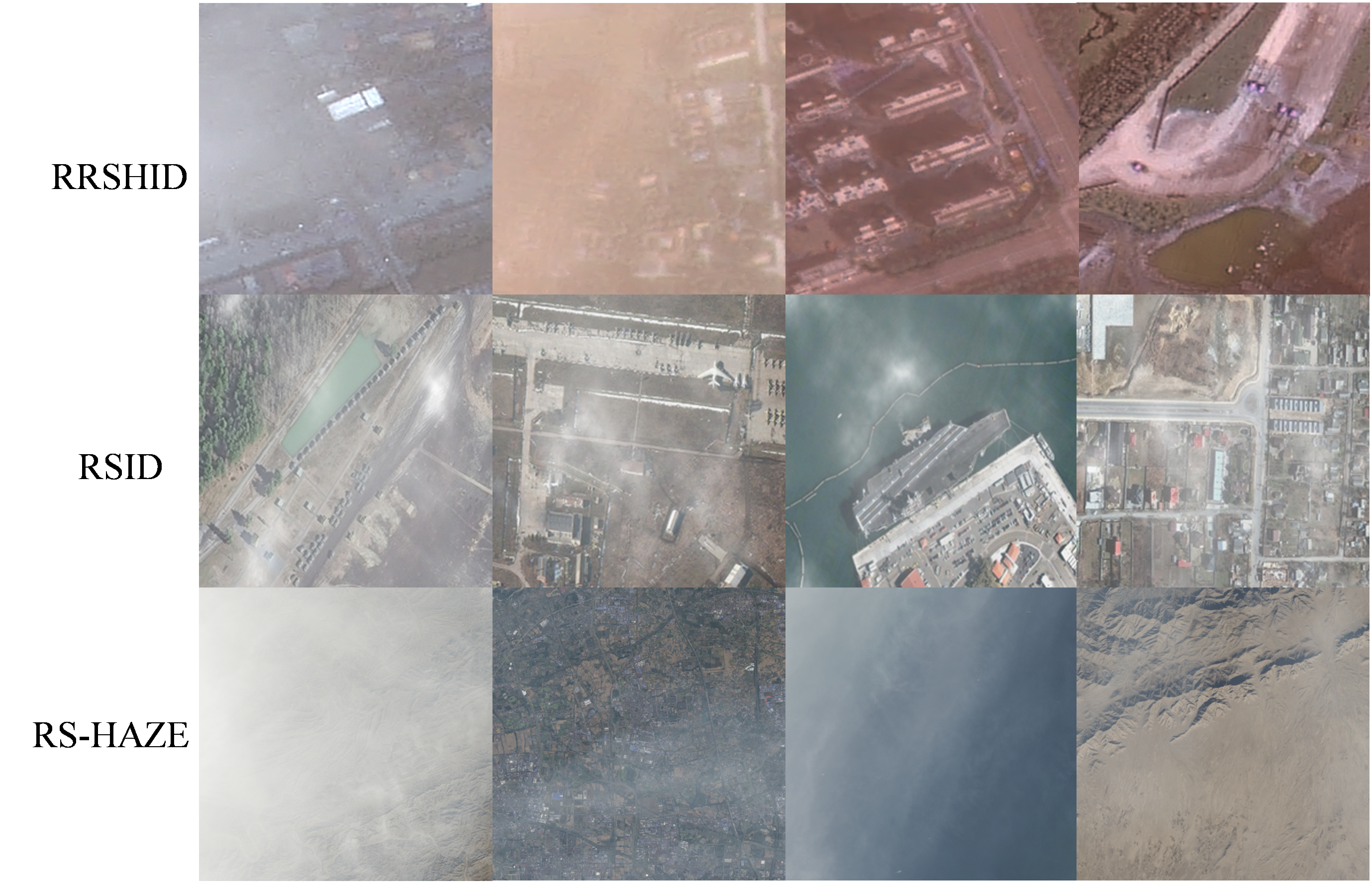}\vspace{-1mm}
	\caption{Sample of the RRSHID that we proposed. Unlike synthetic datasets (e.g., RSID, RS-HAZE), RRSHID demonstrates authentic characteristics in haze density distribution, land cover diversity, and color distortion.}
	\label{fig:sample}
\end{figure}

These properties, as depicted in Fig.~\ref{fig:sample}, render RRSHID a crucial resource for developing robust RSID algorithms, assessing haze impacts on applications such as object detection~\cite{lu2025lwganet,lu2025legnet} and change detection~\cite{wang2024attentionaware,liu2025commonality}, and deepening insights into atmospheric effects on satellite imagery.	 

\section{Experiments}
\subsection{Experimental Settings}

\subsubsection{Datasets}
MCAF-Net is evaluated on the RRSHID dataset and five publicly available benchmarks: RSID \cite{chi2023trinity}, RICE1 \cite{lin2019remote}, RICE2 \cite{lin2019remote}, StateHaze1K-thick \cite{huang2020single}, and HRSI100. These datasets are selected for their diverse characteristics:  

\textbf{RSID}: Includes 1000 hazy and haze-free image pairs, with 900 used for training and 100 for testing.  

\textbf{RICE1}: Comprises 500 cloudy and cloud-free pairs, split into 400 for training and 100 for testing.  

\textbf{RICE2}: Features 736 pairs with complex thick clouds and shadows, divided into 588 for training and 147 for testing.  

\textbf{StateHaze1K-thick}: A subset of 400 hazy pairs, allocated as 320 for training, 45 for testing, and 35 for validation.

\textbf{HRSI100}: Contains 100 real-world RS hazy images cropped to $256\times256$ resolution from the HRSI \cite{liu2024oriented} dataset without ground truth, specifically designed to evaluate generalization capability in real-world scenarios.

The RRSHID dataset, designed to reflect real-world RS challenges, complements these synthetic and real-world benchmarks to comprehensively assess generalization performance.

\subsubsection{Implementation Details}
All models trained for 300 epochs on a single NVIDIA Tesla P100-PCIE GPU. We configure the patch size to $256\times256$ and the batch size to 8. The Adam optimizer is utilized with $\beta_1 = 0.9$ and $\beta_2 = 0.999$, paired with an initial learning rate of 0.001, which decays to $1\times10^{-8}$ via a cosine annealing scheduler. The network architecture comprises five stages, with embedding dimensions of [24, 48, 96, 48, 24] and corresponding depths of [8, 8, 16, 8, 8], optimizing both capacity and efficiency for RSID.

\subsubsection{Evaluation Metrics}
To assess the performance on supervised hazy datasets, we employed four metrics: PSNR, Structural Similarity Index Measure (SSIM), Mean Squared Error (MSE), and Learned Perceptual Image Patch Similarity (LPIPS). Higher PSNR and SSIM values signify improved image quality, while lower MSE and LPIPS indicates enhanced dehazing efficacy. To further validate generalization capability in real-world scenarios, we additionally utilized Natural Image Quality Evaluator (NIQE) \cite{mittal2012making} and Integrated Local NIQE (IL-NIQE) \cite{zhang2015feature} metrics, where lower values indicate better perceptual quality without requiring reference images.

\subsection{Experimental Results}
\begin{table*}[!t]
	\centering
	\scriptsize
	\renewcommand{\arraystretch}{1.2}
	\setlength{\tabcolsep}{4.25pt}
	\caption{Performance comparison of different methods on RRSHID dataset across various haze levels. Values in \red{red} and \blue{blue} denote the best and second-best performance, respectively.}\vspace{-1mm}
	\begin{tabular}{lcccccccccccccccc} \toprule
		\multirow{3}{*}{Method} & \multicolumn{4}{c}{RRSHID-thin} & \multicolumn{4}{c}{RRSHID-moderate} & \multicolumn{4}{c}{RRSHID-thick} & \multicolumn{4}{c}{RRSHID-average} \\
		\cmidrule(lr){2-5} \cmidrule(lr){6-9} \cmidrule(lr){10-13} \cmidrule(l){14-17}
		& PSNR & SSIM & MSE & LPIPS & PSNR & SSIM & MSE & LPIPS & PSNR & SSIM & MSE & LPIPS & PSNR & SSIM & MSE & LPIPS \\
		\midrule
		DCP\cite{he2010single} & 18.46 & 0.4564 & 0.0192 & 0.4851 & 17.80 & 0.4856 & 0.0238 & 0.4700 & 18.39 & 0.4843 & 0.0208 & 0.4996 & 18.22 & 0.4754 & 0.0213 & 0.4849 \\
		FFA-Net \cite{qin2020ffa} & 17.08 & 0.4452 & 0.0326 & 0.5761 & 17.40 & 0.5385 & 0.0346 & 0.5450 & 16.71 & 0.4792 & 0.0377 & 0.5573 & 17.06 & 0.4876 & 0.0350 & 0.5595 \\
		GridDehazeNet\cite{liu2019griddehazenet} & 22.77 & 0.6145 & 0.0069 & \blue{0.4123} & 22.62 & 0.6468 & 0.0083 &\blue{ 0.3833} & 23.96 & 0.7112 & 0.0061 & \blue{0.3947} & 23.12 & \blue{0.6575} & 0.0071 & \blue{0.3968} \\
		4KDehazing\cite{xiao2024single} & \blue{22.83} & \blue{0.6177} & \blue{0.0063} & 0.4352 & 22.47 & \blue{0.6505} & 0.0083 & 0.4590 & 22.55 & 0.6912 & 0.0099 & 0.4754 & 22.62 & 0.6531 & 0.0082 & 0.4565 \\
		SCAnet\cite{guo2023scanet} & 18.37 & 0.4718 & 0.0200 & 0.4827 & 18.11 & 0.0538 & 0.0210 & 0.4962 & 19.07 & 0.5966 & 0.0180 & 0.4734 & 18.52 & 0.3741 & 0.0195 & 0.4841 \\
		Trinity-Net\cite{chi2023trinity} & 20.51 & 0.5728 & 0.0120 & 0.4578 & 22.46 & 0.5728 & 0.0085 & 0.4314 & 24.11 & 0.7234 & 0.0058 & 0.4206 & 22.36 & 0.6230 & \blue{0.0060} & 0.4366 \\
		DehazeFormer\cite{song2023vision} & 22.74 & 0.6005 & 0.0071 & 0.4609 & \blue{23.06} & 0.6137 & 0.0076 & 0.4495 & \blue{24.69} & \blue{0.7143} & \blue{0.0051} & 0.4377 & \blue{23.50} & 0.6428 & 0.0066 & 0.4494 \\
		PCSformer\cite{zhang2024proxy} & 21.83 & 0.5427 & 0.0088 & 0.4963 & 22.09 & 0.5984 & 0.0092 & 0.5021 & 23.71 & 0.6547 & 0.0055 & 0.4996 & 22.54 & 0.5996 & 0.0070 & 0.4993 \\
        PhDnet\cite{LIHE2024102277} & 22.64 & 0.6054 & 0.0072 & 0.4658 & 22.92 & 0.6448 & \blue{0.0075} & 0.4019 & 24.28 & 0.6996 & 0.0053 & 0.3993 & 23.28 & 0.6499 & 0.0067 & 0.4223 \\
		\rowcolor{gray!30}\textbf{MCAF-Net} & \red{23.32} & \red{0.6236} & \red{0.0059} & \red{0.4023} & \red{23.60} & \red{0.6583} & \red{0.0063} & \red{0.3799} & \red{25.40} & \red{0.7221} & \red{0.0040} & \red{0.3942} & \red{24.11} & \red{0.6680} & \red{0.0054} & \red{0.3921} \\
		\bottomrule
	\end{tabular}
	\label{tab:performance_comparison1}
\end{table*}

\begin{figure*}[t]
	\includegraphics[width=1\linewidth]{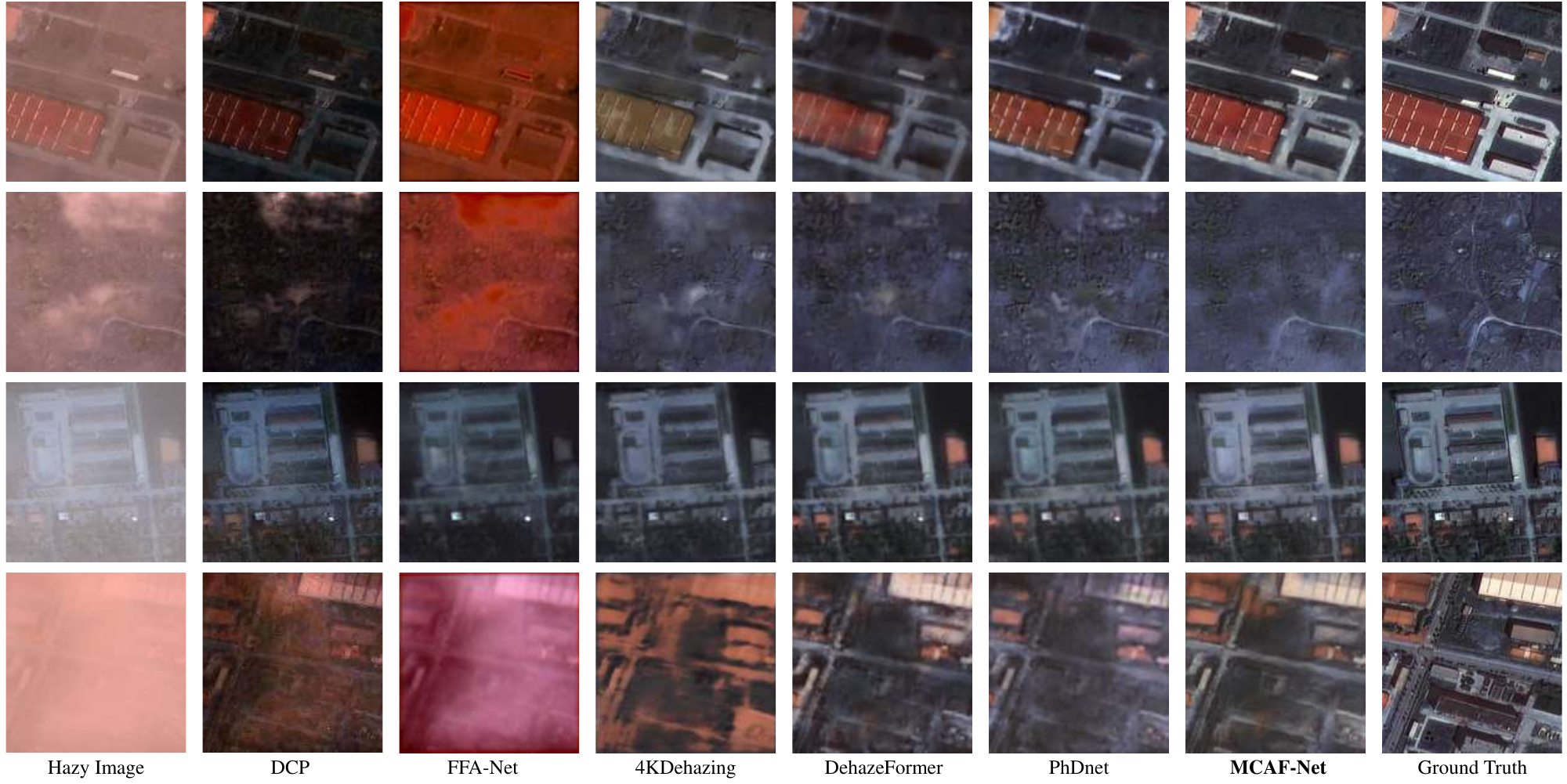}\vspace{-1mm}
	\caption{Visual samples generated by the evaluated methods on the RRSHID dataset. Note that beneath each hazy or restored image, we present the corresponding error maps depicting discrepancies with the ground truth.}
	\label{fig:real}
	\vspace{-2mm}
\end{figure*}

\subsubsection{\textbf{Real-World RSID}}
To evaluate MCAF-Net's effectiveness in real-world RSID, we conducted comprehensive experiments on our RRSHID dataset. This dataset, specifically curated to represent challenging real-world scenarios, served as a robust benchmark for assessing dehazing algorithms in RS imagery.

Table \ref{tab:performance_comparison1} presents a detailed performance comparison of the proposed MCAF-Net against several SOTA dehazing methods on the RRSHID dataset, which encompasses real-world RS images under varying haze conditions: thin, moderate, and thick. The results, as shown in the table, indicate that MCAF-Net consistently achieves the highest performance across all haze levels and metrics, marked in \red{red}, with the second-best performances highlighted in \blue{blue}.

\textbf{RRSHID-thin}: MCAF-Net achieves a PSNR of 23.32, an SSIM of 0.6236, an MSE of 0.0059 and a LPIPS of 0.4023, surpassing all competing methods. The second-best performer, 4KDehazing \cite{xiao2024single}, records a PSNR of 22.83 and an SSIM of 0.6177, trailing MCAF-Net by 0.49 in PSNR and 0.0059 in SSIM, with a higher MSE of 0.0063 and LPIPS of 0.4352.

\textbf{RRSHID-moderate}: MCAF-Net excels with a PSNR of 23.60, an SSIM of 0.6583, an MSE of 0.0063 and a LPIPS of 0.3799. In comparison, DehazeFormer \cite{song2023vision}, the second-best method, achieves a PSNR of 23.06 and an SSIM of 0.6137, falling short by 0.54 in PSNR and 0.0446 in SSIM, with an MSE of 0.0076 and a LPIPS of 0.4495.

\textbf{RRSHID-thick}: MCAF-Net demonstrates superior performance with a PSNR of 25.40, an SSIM of 0.7221, an MSE of 0.0040 and a LPIPS of 0.3942. DehazeFormer obtains a PSNR of 24.69 and an SSIM of 0.7143, lagging by 0.71 in PSNR and 0.0078 in SSIM, with an MSE of 0.0051 and a LPIPS of 0.4377.

\textbf{RRSHID-average}: Across all haze levels, MCAF-Net maintains its lead with an average PSNR of 24.11, an SSIM of 0.6680, an MSE of 0.0054 and a LPIPS of 0.3921. The second-best results vary by metric: DehazeFormer achieves a PSNR of 23.50 (0.61 lower than MCAF-Net), GridDehazeNet \cite{liu2019griddehazenet} records an SSIM of 0.6575 (0.0105 lower) and a LPIPS of 0.3968 (0.0047 higher), and Trinity-Net \cite{chi2023trinity} yields an MSE of 0.0060 (0.0006 higher).

The consistent superiority of MCAF-Net across all haze conditions and evaluation metrics underscores its robustness and effectiveness in real-world RSID. Notably, the performance gap is most pronounced under thick haze conditions, where MCAF-Net achieves significant improvements in PSNR (0.71 higher than DehazeFormer) and MSE (0.0011 lower), indicating its capability to recover fine details and structural integrity in heavily degraded images. 

Compared to established methods such as DCP \cite{he2010single}, FFA-Net \cite{qin2020ffa}, SCAnet \cite{guo2023scanet} and PCSformer \cite{zhang2024proxy}, MCAF-Net demonstrates substantial improvements, especially in SSIM and MSE, reflecting its enhanced ability to preserve structural fidelity and minimize reconstruction errors. Even against more recent approaches such as GridDehazeNet, 4KDehazing, Trinity-Net, DehazeFormer and PhDnet\cite{LIHE2024102277}, MCAF-Net stands out as a superior solution, reinforcing its position as a SOTA method for real-world RSID.

To provide a more comprehensive assessment beyond numerical metrics, we included visual comparisons of the dehazing results in Fig. \ref{fig:real}. These qualitative examples offered insight into the perceptual improvements achieved by our method. Qualitative results further validate the quantitative findings, showcasing our algorithm's ability to effectively remove haze while preserving crucial image details and enhancing overall visual clarity, while FFA-Net and 4KDehazing often leave residual haze or introduce artifacts. Despite MCAF-Net's overall effectiveness across various haze conditions, it still faces challenges in extremely dense haze scenarios. When visibility is severely compromised, the model struggles to recover fine-grained details and occasionally produces artifacts in regions with complex textures. As demonstrated in the last row of Fig. \ref{fig:real}, our method shows performance limitations when dealing with intricate structures such as dense urban areas or complex vegetation patterns.

In summary, the experimental results on the RRSHID dataset validate the efficacy of MCAF-Net for real-world RSID. Its consistent outperformance across diverse haze levels and metrics establishes MCAF-Net as a leading approach in this domain, offering a robust and reliable solution for enhancing the quality of haze-degraded RS imagery.

\subsubsection{\textbf{Synthetic RSID}}

\begin{table*}[!t]
		\centering	\scriptsize
		\renewcommand{\arraystretch}{1.2}
		\setlength{\tabcolsep}{4.25pt}
	\caption{Performance comparison of different methods on multiple haze datasets. Values in \red{red} and \blue{blue} denote the best and second-best performance, respectively.}\vspace{-1mm}
	\begin{tabular}{lcccccccccccccccc}
		\toprule
		\multirow{3}{*}{Method} & \multicolumn{4}{c}{RSID} & \multicolumn{4}{c}{RICE1} & \multicolumn{4}{c}{RICE2} & \multicolumn{4}{c}{Statehaze1k-thick} \\
		\cmidrule(lr){2-5} \cmidrule(lr){6-9} \cmidrule(lr){10-13} \cmidrule(l){14-17}
		& PSNR & SSIM & MSE & LPIPS & PSNR & SSIM & MSE & LPIPS & PSNR & SSIM & MSE & LPIPS & PSNR & SSIM & MSE & LPIPS \\
		\midrule
		DCP\cite{he2010single} & 13.87 & 0.6892 & 0.0538 & 0.2615 & 15.68 & 0.6860 & 0.0491 & 0.3766 & 14.10 & 0.3898 & 0.0656 & 0.6234 & 17.87 & 0.8481 & 0.0169 & 0.1879 \\
		FFA-Net\cite{qin2020ffa} & 18.31 & 0.8582 & 0.0204 & 0.1449 & 23.73 & 0.9068 & 0.0096 & 0.1885 & 17.77 & 0.6261 & 0.0323 & 0.5378 & 19.45 & 0.9023 & 0.0117 & 0.2357 \\
		GridDehazeNet\cite{liu2019griddehazenet} & 23.50 & 0.9383 & 0.0080 & 0.0541 & 33.45 & 0.9766 & 0.0010 & \blue{0.0405} & 31.54 & 0.8839 & 0.0014 & \red{0.1942} & 20.51 & 0.9097 & 0.0089 &\red{ 0.1361} \\
		4KDehazing\cite{xiao2024single} & 23.61 & 0.9415 & 0.0053 & 0.0693 & 27.54 & 0.9425 & 0.0024 & 0.0967 & 25.21 & 0.8604 & 0.0087 & 0.3704 & 20.75 & 0.7696 & 0.0085 & 0.1674 \\
		Trinity-Net\cite{chi2023trinity} & 23.60 & 0.9322 & 0.0060 & 0.0692 & 23.46 & 0.8796 & 0.0100 & 0.0512 & 18.81 & 0.7781 & 0.0200 & 0.2287 & 20.43 & 0.8056 & 0.0086 & 0.1764 \\
		PSMB-Net\cite{sun2023partial} & \blue{25.64} & \blue{0.9447} & \red{0.0030} & \blue{0.0537} & 31.32 & 0.9444 & 0.0006 & 0.0428 & 30.21 & 0.8756 & 0.0010 & 0.1983 & 21.55 & 0.8489 & 0.0084 & 0.1642 \\
		DehazeFormer\cite{song2023vision} & 25.04 & 0.9393 & 0.0040 & 0.0557 & \red{36.15} & \blue{0.9794} & \blue{0.0005} & 0.0460 & \blue{34.54} & 0.8875 & \blue{0.0010} & 0.2035 & \blue{22.02} & \blue{0.9306} & \blue{0.0063} & 0.1707 \\
		PCSformer\cite{zhang2024proxy} & 23.03 & 0.9159 & 0.0063 & 0.0718 & 35.17 & 0.9564 & 0.0006 & 0.0421 & 33.80 & 0.8781 & 0.0011 & 0.2418 & 20.20 & 0.8125 & 0.0097 & 0.1734 \\
        PhDnet\cite{LIHE2024102277} & 25.15 & 0.9406 & 0.0040 & 0.0612 & 35.50 & 0.9578 & 0.0005 & 0.0464 & 33.44 & \blue{0.8877} & 0.0011 & 0.2277 & 21.99 & 0.8294 & 0.0064 & 0.1721 \\
		\rowcolor{gray!30}\textbf{MCAF-Net} & \red{25.89} & \red{0.9531} & \blue{0.0032} & \red{0.0421} & \blue{35.79} & \red{0.9814} & \red{0.0005} & \red{0.0383} & \red{34.59} & \red{0.8948} & \red{0.0009} & \blue{0.1964} & \red{22.16} & \red{0.9308} & \red{0.0062} & \blue{0.1453} \\
		\bottomrule
	\end{tabular}
	\label{tab:performance_comparison2}
	\vspace{-1mm}
\end{table*}

\begin{figure*}[t]
	\centering	\scriptsize
	\includegraphics[width=1\linewidth]{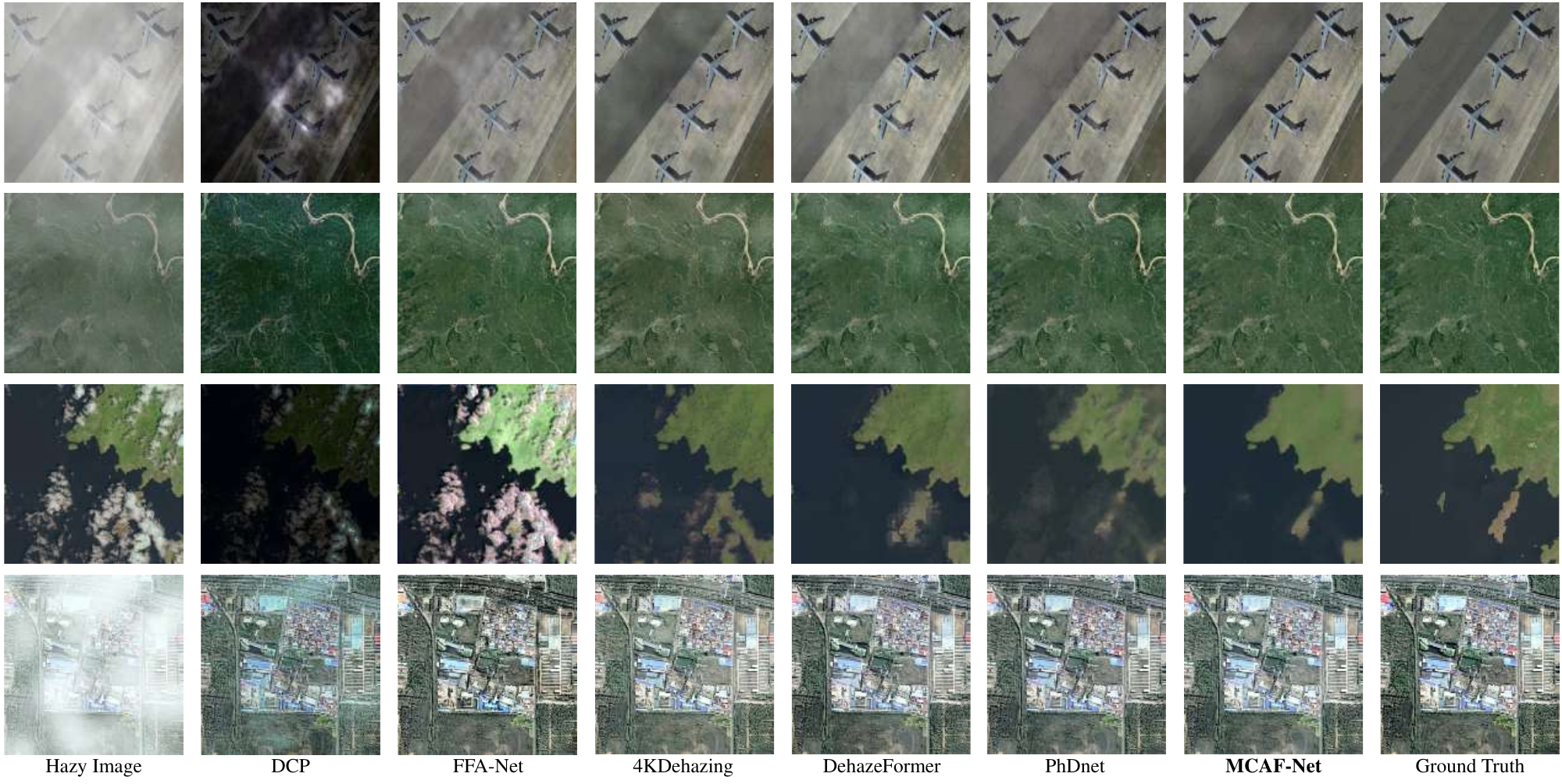}\vspace{-1mm}
	\caption{Visual samples generated by the evaluated methods on RSID \cite{chi2023trinity}, RICE1\cite{lin2019remote}, RICE2\cite{lin2019remote}, and Statehaze1k-thick\cite{huang2020single} datasets.}
	\label{fig:sys}
	\vspace{-2mm}
\end{figure*}

Our method demonstrated exceptional performance in real-world RSID. To further evaluate its generalization capability, we applied the proposed approach to several publicly available synthetic RSID datasets. The methods include DCP \cite{he2010single}, FFA-Net \cite{qin2020ffa}, GridDehazeNet \cite{liu2019griddehazenet}, 4KDehazing \cite{xiao2024single}, Trinity-Net \cite{chi2023trinity}, PSMB-Net \cite{sun2023partial}, DehazeFormer \cite{song2023vision}, PCSformer \cite{zhang2024proxy}, PhDnet\cite{LIHE2024102277} and our proposed MCAF-Net.

Theoretically, the synthetic data exhibited certain scene differences compared to the real data, which could potentially lead to suboptimal results on some public synthetic RS datasets. However, our experimental results were highly encouraging. Our method not only performed admirably on these synthetic datasets but also provided strong evidence of its robust generalization ability and resilience.

We conducted comprehensive comparative analyses across multiple SOTA methods and datasets, including RSID \cite{chi2023trinity}, RICE1 \cite{lin2019remote}, RICE2 \cite{lin2019remote}, and state1k-thick \cite{huang2020single}. 

Experimental results, presented in Table \ref{tab:performance_comparison2}, clearly demonstrated that MCAF-Net outperformed nearly all evaluated approaches in terms of quantitative performance on synthetic RS hazy degradation datasets. Notably, our method exhibited particularly high performance in terms of SSIM scores, indicating its superiority in preserving structural information.

Visual assessments of our method alongside other evaluated approaches were illustrated in Fig. \ref{fig:sys}, further confirming the superiority of our model. These visual comparisons highlighted our method's ability to effectively remove haze while preserving fine details and enhancing overall image quality. 

The consistent superior performance across both real-world and synthetic datasets reinforced the robustness and versatility of our proposed approach, positioning it as a promising solution for a wide range of RSID scenarios.

\subsubsection{\textbf{The generalization verification in Real-World RSID}}

\begin{figure*}[t]
	\includegraphics[width=1\linewidth]{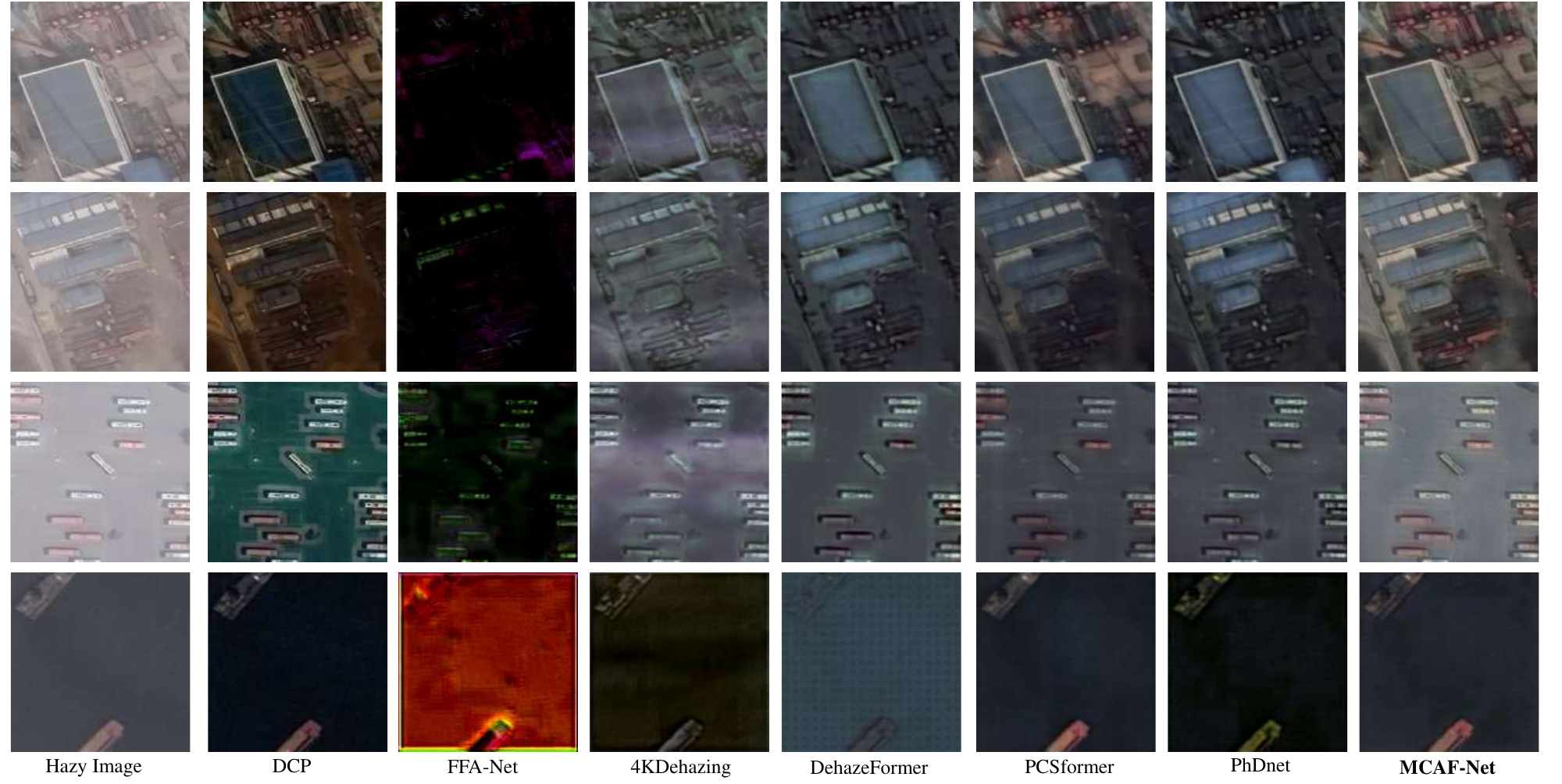}\vspace{-2mm}
	\caption{Visual samples generated by the evaluated methods on the HRSI100 dataset. Our MCAF-Net successfully eliminates haze and better restores image
details, outperforming other evaluated methods.}
	\label{fig:generation}
	\vspace{-2mm}
\end{figure*}

\begin{table}[!t]
		\centering	\scriptsize
		\renewcommand{\arraystretch}{1.2}
	\setlength{\tabcolsep}{24pt}
	\caption{Generalization comparison of state-of-the-art methods}\vspace{-1mm}
	\label{tab:generation comparison}
	\begin{tabular}{lcc}
		\toprule
		Method & NIQE & IL-NIQE \\ 
		\midrule
		DCP \cite{he2010single}             & 6.618   & \blue{65.79}   \\
		FFA-Net \cite{qin2020ffa}        & 18.609  & 120.47  \\
		4KDehazing \cite{xiao2024single}     & 9.207   & 91.79   \\
		DehazeFormer \cite{song2023vision}   & 16.282  & 131.61  \\
		PCSformer \cite{zhang2024proxy}      & \blue{6.105}   & 85.04   \\
            PhDnet \cite{LIHE2024102277}         & 9.684   & 97.12   \\
	\rowcolor{gray!30}	\textbf{MCAF-Net} & \red{5.607} & \red{58.32} \\
		\bottomrule
	\end{tabular}
	\vspace{-2mm}
\end{table}

To further validate the generalization capability of our proposed method and dataset in real-world scenarios, we conducted additional experiments on the HRSI100 dataset. For fair comparison, all methods were evaluated using models trained on the RRSHID-thick dataset without any fine-tuning. Table ~\ref{tab:generation comparison} presents the quantitative results, showing that our method achieves the best NIQE and IL-NIQE scores (5.607 and 58.32 respectively), significantly outperforming other comparison methods. This demonstrates superior generalization ability on unseen real-world hazy scenes. Fig. \ref{fig:generation} provides visual comparisons that further support our findings - our method not only effectively removes haze but also better preserves image details. These results confirm the effectiveness and generalization capability of our proposed method and dataset in real-world RS applications, which is crucial for practical deployment.

\begin{table}[!t]
	\centering	\scriptsize
	\renewcommand{\arraystretch}{1.2}
	\setlength{\tabcolsep}{9pt}
	\caption{Ablation study of different components.}\vspace{-1mm}
	\begin{tabular}{lccccc}
		\toprule
		Model & PSNR & SSIM & MSE & \#Param & FLOPs \\
		\midrule
		Baseline & 23.20 & 0.6952 & 0.0047 & 1.202M & 51.28G \\
		+MFIBA & 24.15 & 0.7091 & 0.0053 & 558.1K & 19.82G \\
		+CSAM & \blue{24.61} & \blue{0.7119} & \blue{0.0045} & \blue{582.3K} & \blue{20.15G} \\
		\textbf{+MFAFM} & \red{25.40} & \red{0.7221} & \red{0.0040} & \red{558.1K} & \red{19.82G} \\
		\bottomrule
	\end{tabular}
	\label{tab:module_1}\vspace{-2mm}
\end{table}

\subsection{Ablation Studies}
The effectiveness of the proposed modules is assessed through ablation studies conducted on the RRSHID-thick dataset. These studies serve to analyze the individual contributions of each component within our proposed MCAF-Net and enable comparisons with their similar designs.

\subsubsection{Individual Components} In this study, we employed a baseline architecture with a U-shaped design, utilizing ResNet-Block (RNB) \cite{he2016deep} as the basic learning block. Table ~\ref{tab:module_1} depicts the ablation on individual components. Initially, we replaced the RNB with our MFIBA, resulting in a performance gain of 0.95 dB in PSNR. Remarkably, this improvement was achieved with only 46.43\% \#Param and 38.65\% FLOPs of the baseline model, demonstrating the efficiency and effectiveness of our MFIBA. Subsequently, incorporating our proposed CSAM module, the model achieved an additional performance gain of 0.46 dB, albeit with a slight increase in computational burden (582.3K \#Param and 20.15G FLOPs). This result indicated the capability of our CSAM in improving performance and demonstrated the significance of color correction in RSID. Finally, we integrated our MFAFM, which led to a substantial improvement of 0.79 dB in PSNR while maintaining similar computational efficiency as MFIBA (558.1K \#Param and 19.82G FLOPs). In conclusion, the ablation studies on individual components demonstrated the significant impact of each proposed component in a positive manner.

\begin{table}[!t]
	\centering	\scriptsize
	\renewcommand{\arraystretch}{1.2}
	\setlength{\tabcolsep}{9pt}
	\caption{Comparison of different basic learning blocks.}\vspace{-1mm}
	\begin{tabular}{lccccc}
		\toprule
		Component & PSNR & SSIM & MSE & \#Param & FLOPs \\
		\midrule
		RNB \cite{he2016deep} & 24.82 & 0.7145 & 0.0048 & 1.15M & 48.56G \\
		FNB \cite{chen2023run} & 24.95 & 0.7168 & 0.0046 & 892.3K & 35.82G \\
		MFIB$\times$1 & 25.13 & 0.7196 & \blue{0.0043} & \red{508.6K} & \red{18.83G} \\
		MFIB$\times$2 & 25.15 & 0.7207 & 0.0044 & \blue{533.3K} & \blue{19.36G} \\
		\textbf{MFIB$\times$3} & \red{25.40} & \red{0.7221} & \red{0.0040} & 558.1K & 19.82G \\
		MFIB$\times$4 & \blue{25.19} & \blue{0.7209} & 0.0044 & 571.3K & 20.02G\\
		\bottomrule
	\end{tabular}
	\label{tab:blocks_1}
	\vspace{-1mm}
\end{table}

\begin{table}[!t]\centering	\scriptsize
	\renewcommand{\arraystretch}{1.2}
	\setlength{\tabcolsep}{6pt}
	\caption{Performance Comparison of Different Components}\vspace{-1mm}
	\label{tab:csam_ablation}
	\begin{tabular}{lccccc}  
		\toprule 
		Component & PSNR & SSIM & CIEDE2000 & \#Param & FLOPs \\
		\midrule
		Baseline          & 23.96  & 0.7112 & 10.38 & \red{534.1K} & \red{18.50G} \\
		+ Attention       & 24.11  & 0.7234 & 7.37  & \blue{542.3K} & \blue{18.75G} \\
		+ Color Calibration & \blue{24.69}  & \blue{0.7143} & \blue{6.27}  & 538.5K & 18.62G \\
		\textbf{+ CSAM}      & \red{25.40} & \red{0.7221} & \red{5.72} & 558.1K & 19.82G \\
		\bottomrule 
	\end{tabular}\vspace{-1mm}
\end{table}

\subsubsection{Basic Learning Blocks} To assess the effectiveness of our proposed MFIBA, we conducted studies comparing its performance with two commonly used learning blocks, namely RNB \cite{he2016deep} and the FasterNetBlock (FNB) \cite{chen2023run}, in the context of RSID. Table~\ref{tab:blocks_1} presented the results of our experiments. Notably, our MFIB$\times$3 outperformed RNB by a margin of 0.58 dB and FNB by a margin of 0.45 dB in terms of PSNR. Furthermore, we investigated the impact of cascading multiple MFIBs. As shown in Table~\ref{tab:blocks_1}, the performance gradually improved as the number of MFIBs increases, with MFIB$\times$3 achieving the best performance while maintaining relatively low computational cost (558.1K \#Param and 19.82G FLOPs). These findings demonstrated the effectiveness of our MFIBA design and its potential for enhancing RSID performance.

\begin{figure}[t]
	\centering	\scriptsize
	\includegraphics[width=1\linewidth]{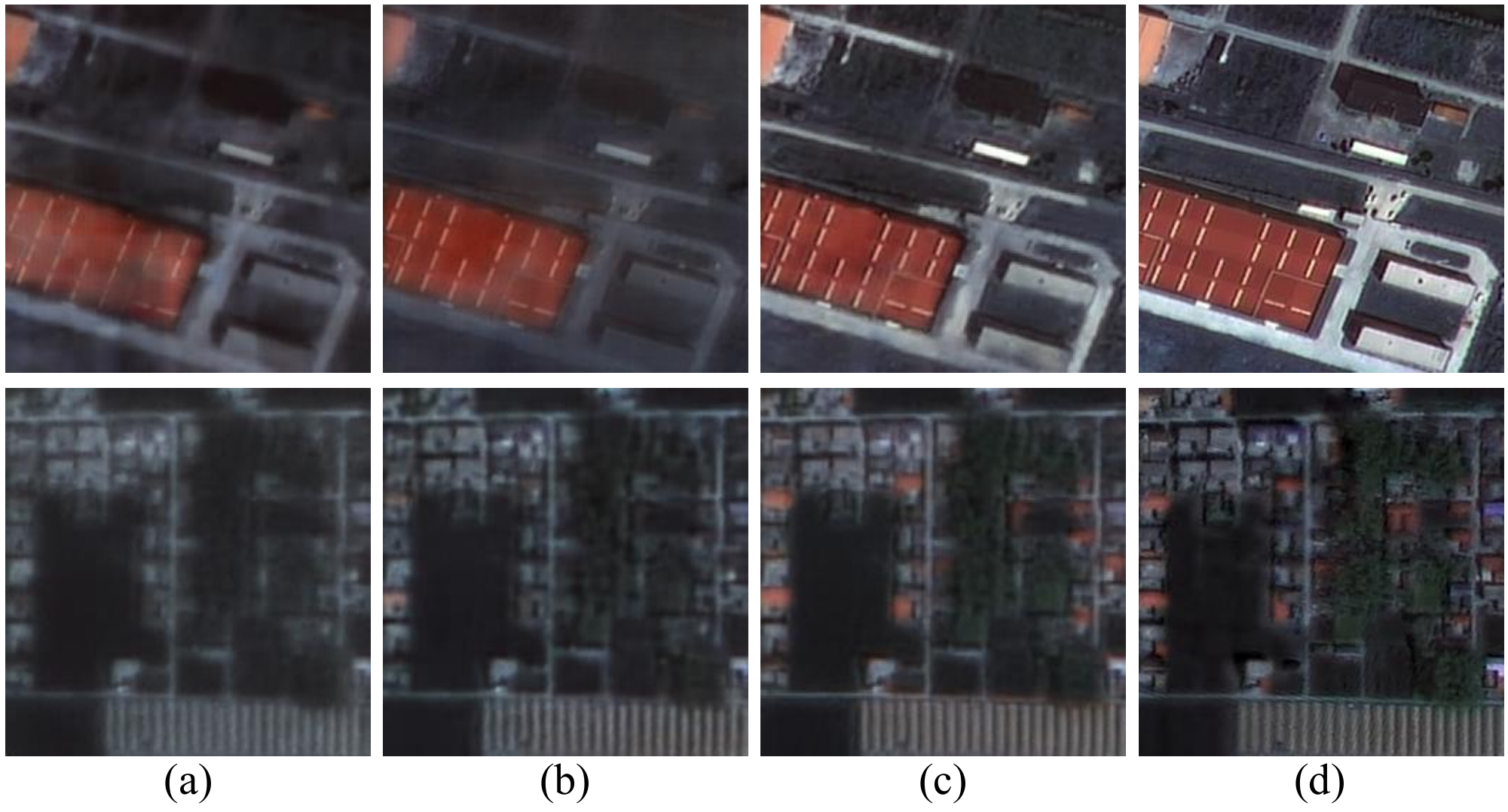}	\vspace{-5mm}
	\caption{Visual comparison of CSAM components: (a)+Attention, (b)+Color Calibration, (c)+CSAM, (d) Ground Truth.}
	\label{fig:component_visualization}
	\vspace{-2mm}
\end{figure}

\subsubsection{Capacity of CSAM} To comprehensively evaluate the effectiveness of the CSAM, we conducted detailed ablation studies focusing on its two core components: the self-supervised color correction mechanism and the attention-guided feature refinement, with the results summarized in Table \ref{tab:csam_ablation}.

The CIEDE2000 \cite{luo2001development} metric, which quantifies perceptual color differences by considering luminance, chroma, and hue variations in the CIELAB color space, demonstrates progressive improvement through our component-wise analysis. When introducing the attention mechanism alone, we observe a moderate PSNR gain of 0.15 dB and SSIM improvement of 0.0122, primarily enhancing structural preservation. However, the substantial CIEDE2000 reduction from 10.38 to 7.37 (-29.0\%) reveals remaining color distortion issues. The color calibration module shows complementary characteristics - it achieves the most significant CIEDE2000 improvement (6.27, -39.6\% from baseline) through learnable color matrix transformation, though with slightly reduced SSIM performance due to enhanced color contrast. Our complete CSAM integration synergistically combines both advantages, delivering SOTA performance with 25.40 dB PSNR (+1.44 dB) and 5.72 CIEDE2000 (-45.0\%), while maintaining practical complexity (558.1K parameters, +4.5\% from baseline).

Fig~\ref{fig:component_visualization} visually validates these findings. The attention-enhanced result (a) preserves building edges but shows yellowish haze residuals. The color-calibrated output (b) exhibits accurate sky color recovery at the expense of vegetation detail. Our full CSAM implementation (c) successfully integrates both capabilities, producing haze-free results with natural chromaticity comparable to GT (d). 

These demonstrate CSAM's unique capacity to bridge the gap between synthetic training and real-world deployment in RSID tasks through physics-guided color correction and data-driven feature refinement.

\begin{table}[!t]
	\centering	\scriptsize
	\renewcommand{\arraystretch}{1.2}
	\setlength{\tabcolsep}{9pt}
	\caption{Performance comparison of different fusion modules}\vspace{-1mm}
	\label{tab:comparison_4}
	\begin{tabular}{lccccc}
		\toprule
		Component & PSNR & SSIM & MSE & \#Param & FLOPs \\
		\midrule
		$1\times 1$ Conv & 25.01 & 0.7128 & 0.0045 & 563.1k & 20.02G \\
		SKFusion \cite{song2023vision} & \blue{25.19} & \blue{0.7132}& \blue{0.0043} & \red{548.5k} & \red{19.42G} \\
		\textbf{MFAFM} & \red{25.40} & \red{0.7221} & \red{0.0040} & \blue{558.1K} & \blue{19.82G} \\
		\bottomrule
	\end{tabular}\vspace{-2mm}
\end{table}

\subsubsection{Adaptive Feature Fusion} To comprehensively validate the efficacy of MFAFM, we performed systematic comparisons with two representative fusion approaches: standard 1$\times$1 convolution and SKFusion \cite{song2023vision}. As quantitatively demonstrated in Table \ref{tab:comparison_4}, the proposed MFAFM achieves marked superiority over conventional convolution operations, achieving a significant gain in PSNR of 0.39 dB. Although SKFusion marginally outperforms baseline convolution (0.12 dB improvement), our method established a new SOTA by further increasing the performance ceiling with an additional 0.21 dB enhancement. This progression not only corroborated the findings in \cite{song2023vision} but more importantly underscored MFAFM's exceptional capability in addressing real-world RSID challenges, confirming MFAFM's effectiveness as a multiscale adaptive paradigm for feature fusion in RSID applications.

\subsubsection{Loss Functions} To investigate the effectiveness of different loss functions, we conducted ablation studies with various loss combinations. Table~\ref{tab:loss} presented the quantitative results. Using only $\mathcal{L}_2$ loss achieved a PSNR of 24.95 dB. When replacing it with perceptual loss, the PSNR decreased to 22.85 dB, but showed better perceptual quality with a SSIM of 0.6873. By combining $\mathcal{L}_2$ and $\mathcal{L}_{perceptual}$ losses, our model achieved the best performance with a PSNR of 25.40 dB and SSIM of 0.7221, demonstrating the complementary effects of these two loss terms in optimizing both pixel-level accuracy and perceptual quality.

\begin{table}[!t]
		\centering	\scriptsize
		\renewcommand{\arraystretch}{1.2}
	\setlength{\tabcolsep}{12pt}
	\caption{Performance comparison of different loss functions.}\vspace{-1mm}
	\begin{tabular}{lccc}
		\toprule
		Loss Function &PSNR & SSIM & MSE \\
		\midrule
		$\mathcal{L}_2$ & \blue{24.95} & \blue{0.7217} & \blue{0.0044} \\[2pt]
		$\mathcal{L}_{perceptual}$ & 22.85 & 0.6873 & 0.0052 \\[2pt]
		$\mathcal{L}_2 + \lambda\mathcal{L}_{perceptual}$ & \red{25.40} & \red{0.7221} & \red{0.0040} \\
		\bottomrule
	\end{tabular}
	\label{tab:loss}
\end{table}

\subsection{Efficiency Analysis}
We conducted a comprehensive comparison of SOTA dehazing methods in Table~\ref{tab:comparison}, evaluating their restoration quality and computational efficiency. MCAF-Net established new benchmarks for RSID by simultaneously achieving SOTA restoration quality (PSNR 25.40 dB, SSIM 0.7221) and unparalleled efficiency (558.1K parameters, 19.82G FLOPs). While Trinity-Net processes images marginally faster (0.0400s vs. 0.0498s), its PSNR lags by 1.29 dB. Compared to DehazeFormer, the prior quality leader, our method reduces the inference time by 43.9\% (0.0831s vs. 0.0498s) while improving the PSNR by 0.71 dB and using 53.7\% fewer parameters. This dual optimization of accuracy and resource efficiency positioned MCAF-Net as the optimal choice for real-time haze removal in resource-constrained environments.
\begin{table}[!t]
		\centering	\scriptsize
		\renewcommand{\arraystretch}{1.2}
	\setlength{\tabcolsep}{6pt}
	\caption{Performance comparison of state-of-the-art methods}\vspace{-1mm}
	\label{tab:comparison}
	\begin{tabular}{lcccccc}
		\toprule
		Method & PSNR & SSIM & MSE & \#Time  & \#Param & FLOPs \\ 
		\midrule
		FFA-Net         & 16.71   & 0.4792  & 0.0377  & 0.1200    & 4.456M   & 287.8G   \\
		GridDehazeNet   & 20.24   & 0.6312  & 0.0096  & 0.0699    & \blue{955.7K}  & 85.72G   \\
		4KDebazing      & 22.55   & 0.6912  & 0.0099  & 0.0549    & 34.55M  & 105.8G   \\
		SCAnet          & 19.07   & 0.5966  & 0.0180  & 0.0700    & 2.39M   & 258.6G   \\
		Trinity-Net     & 24.11   & 0.7103  & 0.0058  & \red{0.0400} & 20.14M & 30.78G  \\
		DehazeFormer    & \blue{24.69}   & \blue{0.7143}  & \blue{0.0051}  & 0.0831    & 1.205M & 39.76G \\
		PCSformer       & 23.71   & 0.6547  & 0.0055  & 0.0600    & 3.73M   & \blue{27.66G}   \\
            PhDnet          & 24.28   & 0.6996  & 0.0053  & 0.0642    & 10.03M   & 33.24G   \\
	\rowcolor{gray!30}	\textbf{MCAF-Net}        & \red{25.40} & \red{0.7221} & \red{0.0040} & \blue{0.0498} & \red{558.1K} & \red{19.82G} \\
		\bottomrule
	\end{tabular}
	\vspace{-2mm}
\end{table}
\section{Conclusion and Future Works}
In this work, we present MCAF-Net, an innovative framework for real-world RSID, designed to overcome the limitations of synthetic-to-real domain gaps through a multi-branch adaptive architecture. Our approach integrates MFIBA, CSAM, and MFAFM to achieve comprehensive feature learning, precise color restoration, and context-aware haze suppression. The introduction of RRSHID dataset—the first large-scale real-world benchmark—provides critical support for modeling complex atmospheric variations and sensor-specific degradations. Extensive validation demonstrates that MCAF-Net not only outperforms existing methods in real-world scenarios but also maintains competitive efficiency with minimal computational overhead, offering practical value for time-sensitive earth observation tasks. While the current framework excels in static haze removal, challenges remain in handling dynamic atmospheric interactions and multi-modal sensor data. Future efforts will focus on integrating temporal modeling and cross-spectral fusion to advance toward all-weather, physics-aware RS restoration systems. Additionally, we recognize that the ultimate value of RSID lies in enhancing the performance of downstream tasks. The RRSHID dataset and MCAF-Net method have the potential to significantly improve applications such as object detection, change detection, and land cover classification under hazy conditions. Haze typically reduces the accuracy of these tasks, especially for small objects and low-contrast features. Our preliminary analysis indicates that using MCAF-Net as a preprocessing step can improve the performance of these tasks. In future work, we plan to construct benchmark datasets with specific task annotations to quantitatively evaluate the impact of dehazing on downstream applications, further validating the practical value of our proposed method in real-world RS applications.

\bibliographystyle{IEEEtran} 
\bibliography{ref}

\begin{thebibliography}{10}
\providecommand{\url}[1]{#1}
\csname url@samestyle\endcsname
\providecommand{\newblock}{\relax}
\providecommand{\bibinfo}[2]{#2}
\providecommand{\BIBentrySTDinterwordspacing}{\spaceskip=0pt\relax}
\providecommand{\BIBentryALTinterwordstretchfactor}{4}
\providecommand{\BIBentryALTinterwordspacing}{\spaceskip=\fontdimen2\font plus
\BIBentryALTinterwordstretchfactor\fontdimen3\font minus
  \fontdimen4\font\relax}
\providecommand{\BIBforeignlanguage}[2]{{%
\expandafter\ifx\csname l@#1\endcsname\relax
\typeout{** WARNING: IEEEtran.bst: No hyphenation pattern has been}%
\typeout{** loaded for the language `#1'. Using the pattern for}%
\typeout{** the default language instead.}%
\else
\language=\csname l@#1\endcsname
\fi
#2}}
\providecommand{\BIBdecl}{\relax}
\BIBdecl

\bibitem{chi2023trinity}
K.~Chi, Y.~Yuan, and Q.~Wang, ``Trinity-net: Gradient-guided swin
  transformer-based remote sensing image dehazing and beyond,'' \emph{IEEE
  Trans. Geosci. Remote Sens.}, vol.~61, pp. 1--14, 2023.

\bibitem{jiang2023dehazing}
B.~Jiang, J.~Wang, Y.~Wu, S.~Wang, J.~Zhang, X.~Chen, Y.~Li, X.~Li, and
  L.~Wang, ``A dehazing method for remote sensing image under nonuniform hazy
  weather based on deep learning network,'' \emph{IEEE Trans. Geosci. Remote
  Sens.}, vol.~61, pp. 1--17, 2023.

\bibitem{gao2023task}
T.~Gao, Z.~Liu, J.~Zhang, G.~Wu, and T.~Chen, ``A task-balanced multiscale
  adaptive fusion network for object detection in remote sensing images,''
  \emph{IEEE Trans. Geosci. Remote Sens.}, vol.~61, pp. 1--15, 2023.

\bibitem{lu2023robust}
W.~Lu, S.-B. Chen, J.~Tang, C.~H.~Q. Ding, and B.~Luo, ``A robust feature
  downsampling module for remote-sensing visual tasks.'' \emph{IEEE Trans.
  Geosci. Remote Sens.}, vol.~61, pp. 1--12, 2023.

\bibitem{lu2024decouplenet}
W.~Lu, S.-B. Chen, Q.-L. Shu, J.~Tang, and B.~Luo, ``Decouplenet: A lightweight
  backbone network with efficient feature decoupling for remote sensing visual
  tasks,'' \emph{IEEE Trans. Geosci. Remote Sens.}, vol.~62, pp. 1--13, 2024.

\bibitem{song2023vision}
Y.~Song, Z.~He, H.~Qian, and X.~Du, ``Vision transformers for single image
  dehazing,'' \emph{IEEE Trans. Image Process.}, vol.~32, pp. 1927--1941, 2023.

\bibitem{he2010single}
K.~He, J.~Sun, and X.~Tang, ``Single image haze removal using dark channel
  prior,'' \emph{IEEE Trans. Pattern Anal. Mach. Intell.}, vol.~33, no.~12, pp.
  2341--2353, 2010.

\bibitem{Valanarasu_2022_CVPR}
J.~M.~J. Valanarasu, R.~Yasarla, and V.~M. Patel, ``Transweather:
  Transformer-based restoration of images degraded by adverse weather
  conditions,'' in \emph{IEEE Conf. Comput. Vis. Pattern Recog.}, June 2022,
  pp. 2353--2363.

\bibitem{xiao2024single}
B.~Xiao, Z.~Zheng, Y.~Zhuang, C.~Lyu, and X.~Jia, ``Single uhd image dehazing
  via interpretable pyramid network,'' \emph{Signal Processing}, vol. 214, p.
  109225, 2024.

\bibitem{lin2019remote}
D.~Lin, G.~Xu, X.~Wang, Y.~Wang, X.~Sun, and K.~Fu, ``A remote sensing image
  dataset for cloud removal,'' \emph{arXiv preprint arXiv:1901.00600}, 2019.

\bibitem{qin2020ffa}
X.~Qin, Z.~Wang, Y.~Bai, X.~Xie, and H.~Jia, ``Ffa-net: Feature fusion
  attention network for single image dehazing,'' \emph{AAAI}, vol.~34, no.~07,
  pp. 11\,908--11\,915, 2020.

\bibitem{narasimhan2002vision}
S.~G. Narasimhan and S.~K. Nayar, ``Vision and the atmosphere,'' \emph{Int. J.
  Comput. Vis.}, vol.~48, pp. 233--254, 2002.

\bibitem{shao2020domain}
Y.~Shao, L.~Li, W.~Ren, C.~Gao, and N.~Sang, ``Domain adaptation for image
  dehazing,'' \emph{IEEE Conf. Comput. Vis. Pattern Recog.}, pp. 2808--2817,
  2020.

\bibitem{qin2018dehazing}
M.~Qin, F.~Xie, W.~Li, Z.~Shi, and H.~Zhang, ``Dehazing for multispectral
  remote sensing images based on a convolutional neural network with the
  residual architecture,'' \emph{IEEE J. Sel. Topics Appl. Earth Observ. Remote
  Sens.}, vol.~11, no.~5, pp. 1645--1655, 2018.

\bibitem{zhang2020multi}
D.~Zhang, J.~Shao, Z.~Liang, X.~Liu, and H.~T. Shen, ``Multi-branch networks
  for video super-resolution with dynamic reconstruction strategy,'' \emph{IEEE
  Trans. Circuit Syst. Video Technol.}, vol.~31, no.~10, pp. 3954--3966, 2020.

\bibitem{zhang2017beyond}
K.~Zhang, W.~Zuo, Y.~Chen, D.~Meng, and L.~Zhang, ``Beyond a gaussian denoiser:
  Residual learning of deep cnn for image denoising,'' \emph{IEEE Trans. Image
  Process.}, vol.~26, no.~7, pp. 3142--3155, 2017.

\bibitem{mao2016image}
X.~Mao, C.~Shen, and Y.-B. Yang, ``Image restoration using very deep
  convolutional encoder-decoder networks with symmetric skip connections,''
  \emph{Adv. Neural Inform. Process. Syst.}, vol.~29, 2016.

\bibitem{zamir2022restormer}
S.~W. Zamir, A.~Arora, S.~Khan, M.~Hayat, F.~S. Khan, and M.-H. Yang,
  ``Restormer: Efficient transformer for high-resolution image restoration,''
  in \emph{IEEE Conf. Comput. Vis. Pattern Recog.}, 2022, pp. 5728--5739.

\bibitem{berman2018single}
D.~Berman, T.~Treibitz, and S.~Avidan, ``Single image dehazing using
  haze-lines,'' \emph{IEEE Trans. Pattern Anal. Mach. Intell.}, vol.~42, no.~3,
  pp. 720--734, 2018.

\bibitem{zhu2015fast}
Q.~Zhu, J.~Mai, and L.~Shao, ``Fast single image haze removal algorithm using
  color attenuation prior,'' in \emph{IEEE Trans. Image Process.}, vol.~24,
  no.~11, 2015, pp. 3522--3533.

\bibitem{berman2016non}
D.~Berman, T.~Treibitz, and S.~Avidan, ``Non-local image dehazing,'' in
  \emph{IEEE Conf. Comput. Vis. Pattern Recog.}, 2016, pp. 1674--1682.

\bibitem{cai2016dehazenet}
B.~Cai, X.~Xu, K.~Jia, C.~Qing, and D.~Tao, ``Dehazenet: An end-to-end system
  for single image haze removal,'' in \emph{IEEE Trans. Image Process.},
  vol.~25, no.~11, 2016, pp. 5187--5198.

\bibitem{ren2016single}
W.~Ren, S.~Liu, H.~Zhang, J.~Pan, X.~Cao, and M.-H. Yang, ``Single image
  dehazing via multi-scale convolutional neural networks,'' in \emph{Eur. Conf.
  Comput. Vis.}\hskip 1em plus 0.5em minus 0.4em\relax Springer, 2016, pp.
  154--169.

\bibitem{li2017aod}
B.~Li, X.~Peng, Z.~Wang, J.~Xu, and D.~Feng, ``Aod-net: All-in-one dehazing
  network,'' in \emph{Int. Conf. Comput. Vis.}, 2017, pp. 4770--4778.

\bibitem{liu2019griddehazenet}
X.~Liu, Y.~Ma, Z.~Shi, and J.~Chen, ``Griddehazenet: Attention-based
  multi-scale network for image dehazing,'' in \emph{Int. Conf. Comput. Vis.},
  2019, pp. 7314--7323.

\bibitem{guo2020rsdehazenet}
J.~Guo, J.~Yang, H.~Yue, H.~Tan, C.~Hou, and K.~Li, ``Rsdehazenet: Dehazing
  network with channel refinement for multispectral remote sensing images,''
  \emph{IEEE Trans. Geosci. Remote Sens.}, vol.~59, no.~3, pp. 2535--2549,
  2020.

\bibitem{fu2024hyperdehazing}
H.~Fu, Z.~Ling, G.~Sun, J.~Ren, A.~Zhang, L.~Zhang, and X.~Jia,
  ``Hyperdehazing: A hyperspectral image dehazing benchmark dataset and a deep
  learning model for haze removal,'' \emph{ISPRS J. Photogramm. Remote Sens.},
  vol. 218, pp. 663--677, 2024.

\bibitem{sun2023partial}
H.~Sun, Z.~Luo, D.~Ren, W.~Hu, B.~Du, W.~Yang, J.~Wan, and L.~Zhang, ``Partial
  siamese with multiscale bi-codec networks for remote sensing image haze
  removal,'' \emph{IEEE Trans. Geosci. Remote Sens.}, vol.~61, pp. 1--16, 2023.

\bibitem{sun2025dynamic}
H.~Sun, S.~Li, B.~Du, L.~Zhang, D.~Ren, and L.~Tong, ``Dynamic-routing
  3d-fusion network for remote sensing image haze removal,'' \emph{IEEE Trans.
  Geosci. Remote Sens.}, 2025.

\bibitem{song2023learning}
T.~Song, S.~Fan, P.~Li, J.~Jin, G.~Jin, and L.~Fan, ``Learning an effective
  transformer for remote sensing satellite image dehazing,'' \emph{IEEE Geosci.
  Remote Sens. Lett.}, vol.~20, pp. 1--5, 2023.

\bibitem{zhang2024proxy}
X.~Zhang, F.~Xie, H.~Ding, S.~Yan, and Z.~Shi, ``Proxy and cross-stripes
  integration transformer for remote sensing image dehazing,'' \emph{IEEE
  Trans. Geosci. Remote Sens.}, 2024.

\bibitem{zhou2024rsdehamba}
H.~Zhou, X.~Wu, H.~Chen, X.~Chen, and X.~He, ``Rsdehamba: lightweight vision
  mamba for remote sensing satellite image dehazing,'' \emph{arXiv preprint
  arXiv:2405.10030}, 2024.

\bibitem{fu2024hdmba}
H.~Fu, G.~Sun, Y.~Li, J.~Ren, A.~Zhang, C.~Jing, and P.~Ghamisi, ``Hdmba:
  hyperspectral remote sensing imagery dehazing with state space model,''
  \emph{arXiv preprint arXiv:2406.05700}, 2024.

\bibitem{guo2023scanet}
Y.~Guo, Y.~Gao, W.~Liu, Y.~Lu, J.~Qu, S.~He, and W.~Ren, ``Scanet: Self-paced
  semi-curricular attention network for non-homogeneous image dehazing,'' in
  \emph{IEEE Conf. Comput. Vis. Pattern Recog.}, 2023, pp. 1885--1894.

\bibitem{li2020semi}
L.~Li, Y.~Dong, W.~Ren, J.~Pan, C.~Gao, N.~Sang, and M.-H. Yang,
  ``Semi-supervised image dehazing,'' \emph{IEEE Trans. Image Process.},
  vol.~29, pp. 2766--2779, 2020.

\bibitem{liang2025image}
Y.~Liang, S.~Li, D.~Cheng, W.~Wang, D.~Li, and J.~Liang, ``Image dehazing via
  self-supervised depth guidance,'' \emph{Pattern Recognition}, vol. 158, p.
  111051, 2025.

\bibitem{lan2025exploiting}
Y.~Lan, Z.~Cui, C.~Liu, J.~Peng, N.~Wang, X.~Luo, and D.~Liu, ``Exploiting
  diffusion prior for real-world image dehazing with unpaired training,'' in
  \emph{AAAI}, vol.~39, no.~4, 2025, pp. 4455--4463.

\bibitem{su2025real}
Y.~Su, N.~Wang, Z.~Cui, Y.~Cai, C.~He, and A.~Li, ``Real scene single image
  dehazing network with multi-prior guidance and domain transfer,'' \emph{IEEE
  Trans. Multimedia}, 2025.

\bibitem{cong2024semi}
X.~Cong, J.~Gui, J.~Zhang, J.~Hou, and H.~Shen, ``A semi-supervised nighttime
  dehazing baseline with spatial-frequency aware and realistic brightness
  constraint,'' in \emph{IEEE Conf. Comput. Vis. Pattern Recog.}, 2024, pp.
  2631--2640.

\bibitem{wu2023maxdehazenet}
H.~Wu, Y.~Qu, J.~Lin, J.~Zhou, and J.~Xiao, ``Maxdehazenet: Multi-scale feature
  aggregation network with maximum flow theory for single image dehazing,''
  \emph{IEEE Trans. Multimedia}, vol.~25, pp. 4876--4889, 2023.

\bibitem{li2019benchmarking}
B.~Li, W.~Ren, D.~Fu, D.~Tao, D.~Feng, W.~Zeng, and Z.~Wang, ``Benchmarking
  single-image dehazing and beyond,'' \emph{IEEE Trans. Image Process.},
  vol.~28, no.~1, pp. 492--505, 2019.

\bibitem{chen2021pre}
H.~Chen, Y.~Wang, T.~Guo, C.~Xu, Y.~Deng, Z.~Liu, S.~Ma, C.~Xu, C.~Xu, and
  W.~Gao, ``Pre-trained image processing transformer,'' in \emph{IEEE Conf.
  Comput. Vis. Pattern Recog.}, 2021, pp. 12\,299--12\,310.

\bibitem{howard2017mobilenets}
A.~G. Howard, M.~Zhu, B.~Chen, D.~Kalenichenko, W.~Wang, T.~Weyand,
  M.~Andreetto, and H.~Adam, ``Mobilenets: Efficient convolutional neural
  networks for mobile vision applications,'' \emph{arXiv preprint
  arXiv:1704.04861}, 2017.

\bibitem{li2022m2scn}
S.~Li, Y.~Zhou, and W.~Xiang, ``M2scn: Multi-model self-correcting network for
  satellite remote sensing single-image dehazing,'' \emph{IEEE Geosci. Remote
  Sens. Lett.}, vol.~20, pp. 1--5, 2022.

\bibitem{li2023efficient}
C.~Li, H.~Yu, S.~Zhou, Z.~Liu, Y.~Guo, X.~Yin, and W.~Zhang, ``Efficient
  dehazing method for outdoor and remote sensing images,'' \emph{IEEE J. Sel.
  Topics Appl. Earth Observ. Remote Sens.}, vol.~16, pp. 4516--4528, 2023.

\bibitem{zheng2022dehaze}
Y.~Zheng, J.~Su, S.~Zhang, M.~Tao, and L.~Wang, ``Dehaze-aggan: Unpaired remote
  sensing image dehazing using enhanced attention-guide generative adversarial
  networks,'' \emph{IEEE Trans. Geosci. Remote Sens.}, vol.~60, pp. 1--13,
  2022.

\bibitem{liu2019single}
Z.~Liu, B.~Xiao, M.~Alrabeiah, K.~Wang, and J.~Chen, ``Single image dehazing
  with a generic model-agnostic convolutional neural network,'' \emph{IEEE
  Sign. Process. Letters}, vol.~26, no.~6, pp. 833--837, 2019.

\bibitem{yang2017rgb}
J.~Yang, Q.~Min, W.~Lu, Y.~Ma, W.~Yao, and T.~Lu, ``An rgb channel operation
  for removal of the difference of atmospheric scattering and its application
  on total sky cloud detection,'' \emph{Atmospheric Measurement Techniques},
  vol.~10, no.~3, pp. 1191--1201, 2017.

\bibitem{ruan2022feature}
Z.~Ruan, Y.~Liu, G.~Wen, J.~Liu, Z.~Hou, and Y.~Zhu, ``Feature distillation
  interaction weighting network for lightweight image super-resolution,''
  \emph{AAAI}, vol.~36, no.~2, pp. 2189--2197, 2022.

\bibitem{xue2023smalltrack}
Y.~Xue, G.~Jin, T.~Shen, L.~Tan, N.~Wang, J.~Gao, and L.~Wang, ``Smalltrack:
  Wavelet pooling and graph enhanced classification for uav small object
  tracking,'' \emph{IEEE Trans. Geosci. Remote Sens.}, vol.~61, pp. 1--15,
  2023.

\bibitem{xue2024consistent}
------, ``Consistent representation mining for multi-drone single object
  tracking,'' \emph{IEEE Trans. Circuit Syst. Video Technol.}, 2024.

\bibitem{gao2023frequency}
T.~Gao, Y.~Wen, K.~Zhang, J.~Zhang, T.~Chen, L.~Liu, and W.~Luo,
  ``Frequency-oriented efficient transformer for all-in-one weather-degraded
  image restoration,'' \emph{IEEE Trans. Circuit Syst. Video Technol.}, 2023.

\bibitem{zhang2020deep}
L.~Zhang, L.~Zhang, and B.~Du, ``Deep learning for remote sensing image
  understanding,'' \emph{Journal of Sensors}, vol. 2016, 2020.

\bibitem{huang2020single}
B.~Huang, L.~Zhi, C.~Yang, F.~Sun, and Y.~Song, ``Single satellite optical
  imagery dehazing using sar image prior based on conditional generative
  adversarial networks,'' in \emph{IEEE Winter Conf. Appl. Comput. Vis.}, 2020,
  pp. 1806--1813.

\bibitem{zhang2022dense}
L.~Zhang and S.~Wang, ``Dense haze removal based on dynamic collaborative
  inference learning for remote sensing images,'' \emph{IEEE Trans. Geosci.
  Remote Sens.}, vol.~60, pp. 1--16, 2022.

\bibitem{liu2024oriented}
B.~Liu, S.-B. Chen, J.-X. Wang, J.~Tang, and B.~Luo, ``An oriented object
  detector for hazy remote sensing images,'' \emph{IEEE Trans. Geosci. Remote
  Sens.}, 2024.

\bibitem{lu2025lwganet}
W.~Lu, S.-B. Chen, C.~H.~Q. Ding, J.~Tang, and B.~Luo, ``Lwganet: A lightweight
  group attention backbone for remote sensing visual tasks,'' \emph{arXiv
  preprint arXiv:2501.10040}, 2025.

\bibitem{lu2025legnet}
W.~Lu, S.-B. Chen, H.-D. Li, Q.-L. Shu, C.~H.~Q. Ding, J.~Tang, and B.~Luo,
  ``Legnet: Lightweight edge-{Gaussian} driven network for low-quality remote
  sensing image object detection,'' \emph{arXiv preprint arXiv:2503.14012},
  2025.

\bibitem{wang2024attentionaware}
L.~Wang, Z.-H. You, W.~Lu, S.-B. Chen, J.~Tang, and B.~Luo, ``Attention-aware
  sobel graph convolutional network for remote sensing image change
  detection,'' \emph{IEEE Trans. Geosci. Remote Sens.}, vol.~62, pp. 1--12,
  2024.

\bibitem{liu2025commonality}
T.~Liu, M.~Zhang, M.~Gong, Q.~Zhang, F.~Jiang, H.~Zheng, and D.~Lu,
  ``Commonality feature representation learning for unsupervised multimodal
  change detection,'' \emph{IEEE Trans. Image Process.}, vol.~34, pp.
  1219--1233, 2025.

\bibitem{mittal2012making}
A.~Mittal, R.~Soundararajan, and A.~C. Bovik, ``Making a “completely blind”
  image quality analyzer,'' \emph{IEEE Sign. Process. Letters}, vol.~20, no.~3,
  pp. 209--212, 2012.

\bibitem{zhang2015feature}
L.~Zhang, L.~Zhang, and A.~C. Bovik, ``A feature-enriched completely blind
  image quality evaluator,'' \emph{IEEE Trans. Image Process.}, vol.~24, no.~8,
  pp. 2579--2591, 2015.

\bibitem{LIHE2024102277}
Z.~Lihe, J.~He, Q.~Yuan, X.~Jin, Y.~Xiao, and L.~Zhang, ``Phdnet: A novel
  physic-aware dehazing network for remote sensing images,'' \emph{Information
  Fusion}, vol. 106, p. 102277, 2024.

\bibitem{he2016deep}
K.~He, X.~Zhang, S.~Ren, and J.~Sun, ``Deep residual learning for image
  recognition,'' in \emph{IEEE Conf. Comput. Vis. Pattern Recog.}, 2016, pp.
  770--778.

\bibitem{chen2023run}
J.~Chen, S.-h. Kao, H.~He, W.~Zhuo, S.~Wen, C.-H. Lee, and S.-H.~G. Chan,
  ``Run, don't walk: chasing higher flops for faster neural networks,'' in
  \emph{IEEE Conf. Comput. Vis. Pattern Recog.}, 2023, pp. 12\,021--12\,031.

\bibitem{luo2001development}
M.~R. Luo, G.~Cui, and B.~Rigg, ``The development of the cie 2000
  colour-difference formula: Ciede2000,'' \emph{Color Research \& Application},
  vol.~26, no.~5, pp. 340--350, 2001.

\end{thebibliography}

\begin{IEEEbiography}[{\includegraphics[width=1.25in,height=1.25in,clip,keepaspectratio]{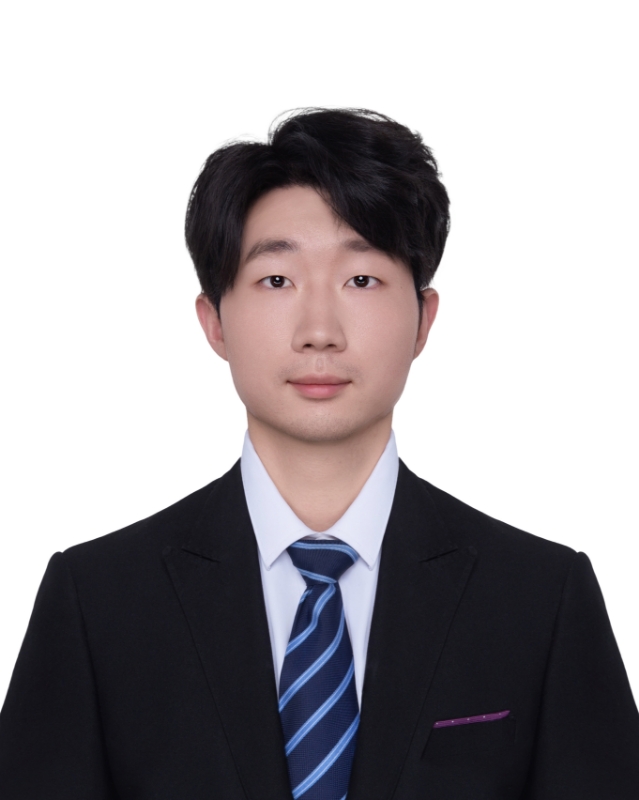}}]{Zeng-Hui Zhu} received the B.S. degree in computer science from Anhui Wenda University of Information Engineering, Hefei, China, in 2022. He is currently pursuing the master's degree in computer science with Anhui University, Hefei. His research interests include machine learning, pattern recognition, computer vision, and remote sensing image dehazing. \end{IEEEbiography} 

\begin{IEEEbiography}[{\includegraphics[width=1.25in,height=1.25in,clip,keepaspectratio]{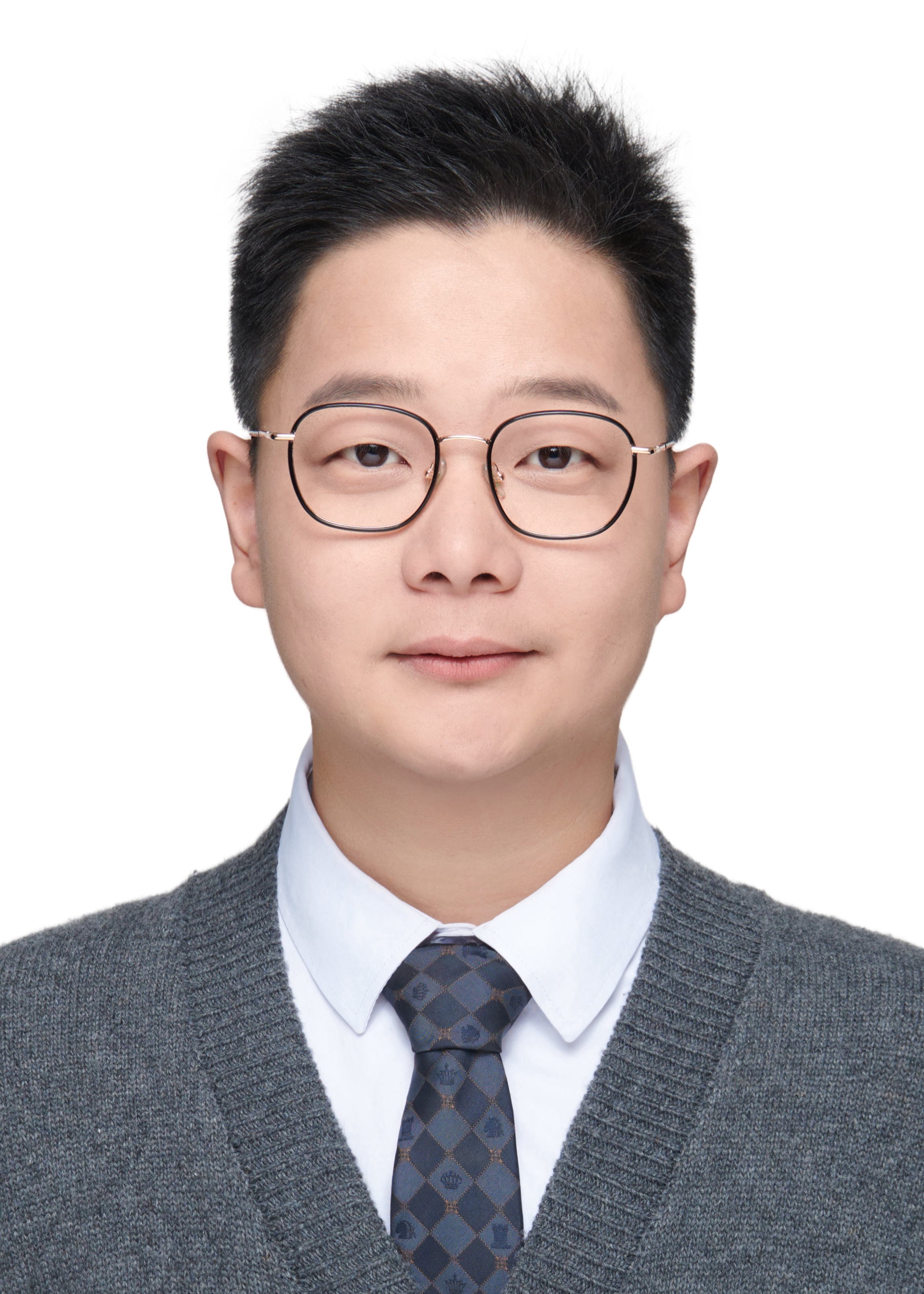}}]{Wei Lu} received the B.Eng. degree from Fuyang Normal University, China in 2020. He is currently pursuing a Ph.D. in computer science from Anhui University, Hefei, China. His current research interests include image processing, efficient AI, low-level visual tasks, and remote sensing vision.  \end{IEEEbiography}

\begin{IEEEbiography}[{\includegraphics[width=1.25in,height=1.25in,clip,keepaspectratio]{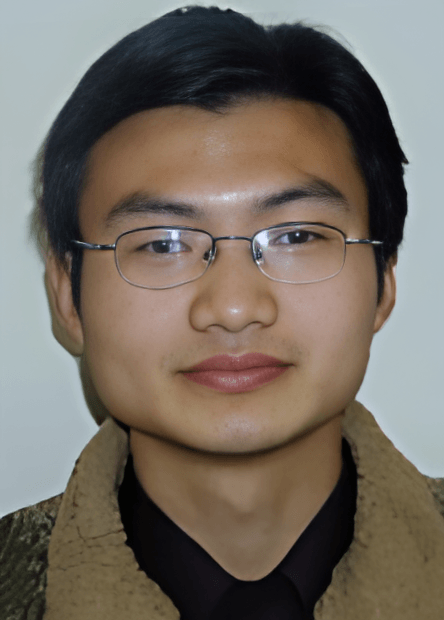}}]{Si-Bao Chen} received the B.S. and M.S. degrees in probability and statistics and the Ph.D. degree in computer science from Anhui University of China in 2000, 2003, and 2006, respectively. From 2006 to 2008, he was a Postdoctoral Researcher at the University of Science and Technology of China. Since 2008, he has been a teacher at Anhui University. From 2014 to 2015, he was a visiting scholar at the University of Texas in Arlington. His current research interests include image processing, pattern recognition, machine learning and computer vision.\end{IEEEbiography}

\begin{IEEEbiography}[{\includegraphics[width=1.0in,height=1.25in,clip,keepaspectratio]{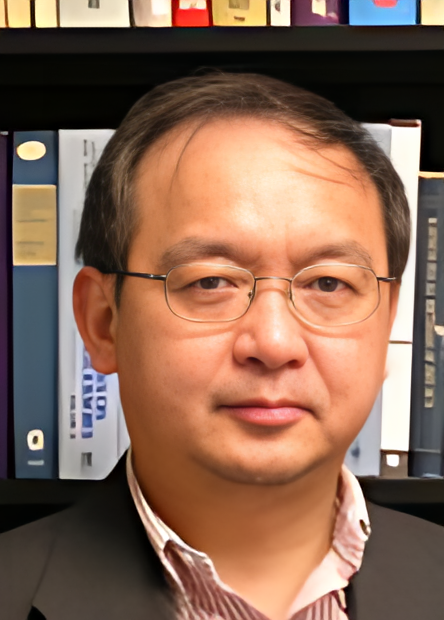}}]{Chris H.Q. Ding} obtained his Ph.D. from Columbia University in 1987. He has held positions at the California Institute of Technology, Jet Propulsion Laboratory, Lawrence Berkeley National Laboratory, University of California, and the University of Texas in Arlington. Since 2021, he has been affiliated with the Chinese University of Hong Kong, Shenzhen. Dr. Ding is an expert in machine learning, data mining, bioinformatics, information retrieval, web link analysis, and high-performance computing. His works were cited over 60,000 times (Google scholar).\end{IEEEbiography} 

\begin{IEEEbiography}[{\includegraphics[width=1.25in,height=1.25in,clip,keepaspectratio]{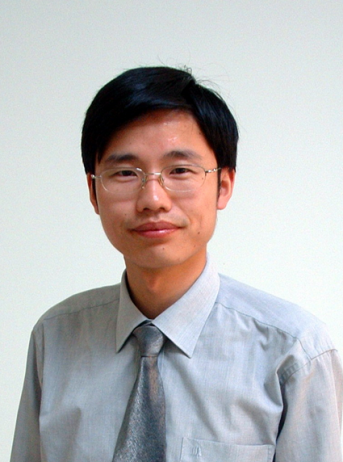}}]{Jin Tang} received the B.Eng. degree in automation and the Ph.D. degree in computer science from Anhui University, Hefei, China, in 1999 and 2007, respectively. He is currently a Professor with the School of Computer Science and Technology, Anhui University. His current research interests include computer vision, pattern recognition, machine learning, and deep learning. His works were cited cited over 10,000 times (Google scholar). \end{IEEEbiography}

\begin{IEEEbiography}[{\includegraphics[width=1.25in,height=1.25in,clip,keepaspectratio]{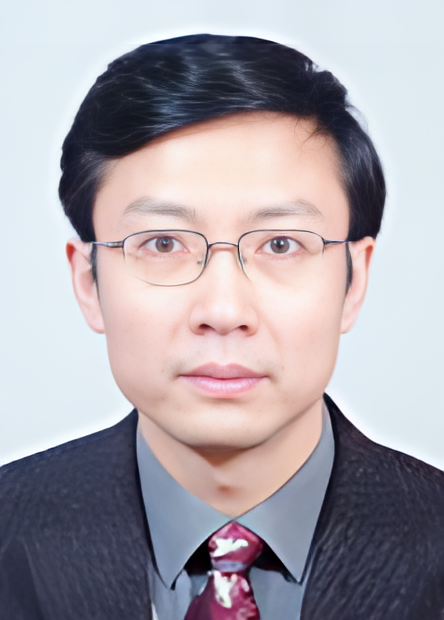}}]{Bin Luo} received the B.Eng. degree in electronics and the M.Eng. degree in computer science from Anhui University, Hefei, China, in 1984 and 1991, respectively, and the Ph.D. degree in computer science from the University of York, York, U.K., in 2002. From 2000 to 2004, he was a Research Associate with the University of York. He is currently a Professor with Anhui University. His current research interests include graph spectral analysis, image and graph matching, statistical pattern recognition, digital watermarking, and information security. His works were cited over 10,000 times (Google scholar).\end{IEEEbiography}

\end{document}